\documentclass{article}
\usepackage{amsthm}
\usepackage{ijcai24}
\usepackage{times}
\usepackage{soul}
\usepackage{url}
\usepackage{natbib}
\usepackage[hidelinks]{hyperref}
\usepackage[utf8]{inputenc}
\usepackage[small]{caption}
\usepackage{graphicx}
\usepackage{subcaption}
\usepackage{amsmath}
\usepackage{thm-restate}
\usepackage{booktabs}
\usepackage{algorithm}
\usepackage{graphicx}
\usepackage[table]{xcolor}
\definecolor{lightgray}{gray}{0.9}
\usepackage{algpseudocode}
\usepackage[switch]{lineno}
\usepackage{amssymb}
\usepackage{multirow}
\usepackage{float}

\newtheorem{definition}{Definition}

\title{Machine Unlearning via Null Space Calibration}

\author{
Huiqiang Chen$^{1,3}$
\footnote{word done while at University of Queensland}
\and
Tianqing Zhu$^{2}$ \footnote{corresponding author}
\and
Xin Yu$^{3}$\And
Wanlei Zhou$^4$
\affiliations
$^1$University of Technology Sydney, NSW, Australia\\
$^2$City University of Macau, Macau, China\\
$^3$University of Queensland, QLD, Australia\\
\emails
huiqiang.chen@student.uts.edu.au,
\{tqzhu, wlzhou\}@cityu.edu.mo,
xin.yu@uq.edu.au
}

\begin{document}
\maketitle
\thispagestyle{plain}
\pagestyle{plain}  
\pagenumbering{arabic}
\begin{abstract}
    Machine unlearning aims to enable models to forget specific data instances when receiving deletion requests. Current research centres on efficient unlearning to erase the influence of data from the model and neglects the subsequent impacts on the remaining data. Consequently, existing unlearning algorithms degrade the model's performance after unlearning, known as \textit{over-unlearning}. This paper addresses this critical yet under-explored issue by introducing machine \underline{U}nlearning via \underline{N}ull \underline{S}pace \underline{C}alibration (UNSC), which can accurately unlearn target samples without over-unlearning. On the contrary, by calibrating the decision space during unlearning, UNSC can significantly improve the model's performance on the remaining samples. In particular, our approach hinges on confining the unlearning process to a specified null space tailored to the remaining samples, which is augmented by strategically pseudo-labeling the unlearning samples. Comparative analyses against several established baselines affirm the superiority of our approach. Code is released at this \href{https://github.com/HQC-ML/Machine-Unlearning-via-Null-Space-Calibration}{URL}.
\end{abstract}

\section{Introduction}
With the rising concerns regarding the privacy of users' data in the AI era, legislative bodies have instituted regulatory measures to safeguard user's data. A notable example is the \textit{Right to Be Forgotten}, which requires service providers to remove users' data from AI models upon receiving deletion requests \citep{regulation2018general}. 
However, removing the trace of particular data from a trained model is challenging. Machine unlearning has emerged as a promising solution, it endeavors to enable models to forget specific data samples, effectively erasing their influence as if they were never trained on these samples. The simplest way is to retrain the model from scratch without unlearning samples. However, it is time-consuming and even impossible in the face of frequent unlearning requests. As such, the research community has been actively developing algorithms to expedite unlearning. 

\begin{figure}[htp]
  \centering
  \includegraphics[width=0.95\linewidth]{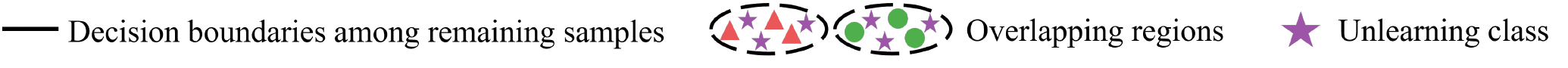}
  \begin{minipage}{.32\linewidth}
    \includegraphics[width=\linewidth]{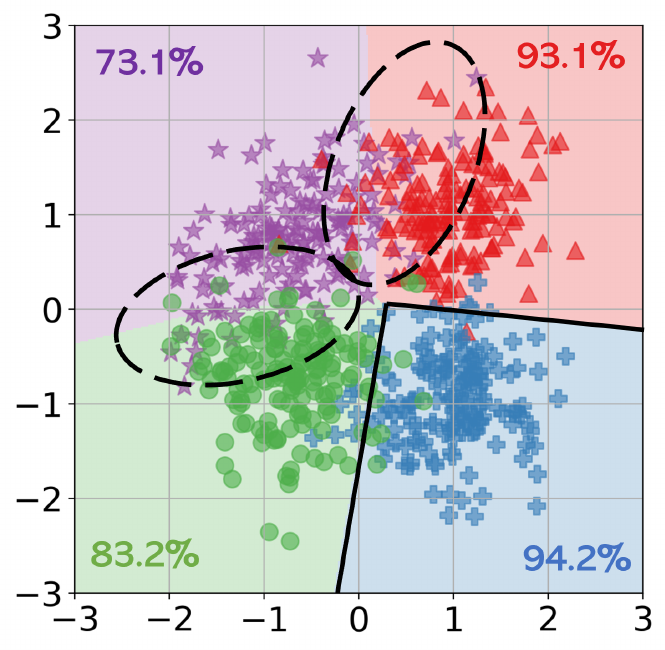}
    \vspace{-1.5em}
    \caption*{(a) Original model}
  \end{minipage}%
  \hfill
  \begin{minipage}{.32\linewidth}
    \includegraphics[width=\linewidth]{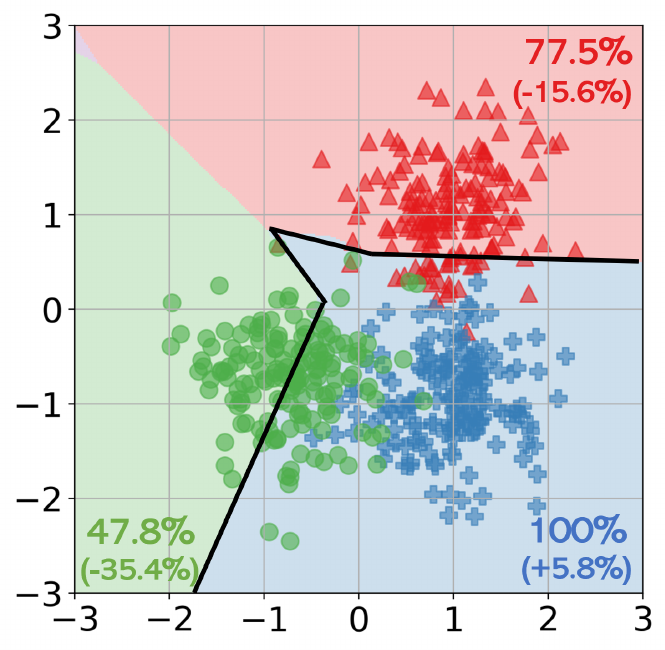}
    \vspace{-1.5em}
    \caption*{(b) Baseline}
  \end{minipage}%
  \hfill
  \begin{minipage}{.32\linewidth}
    \includegraphics[width=\linewidth]{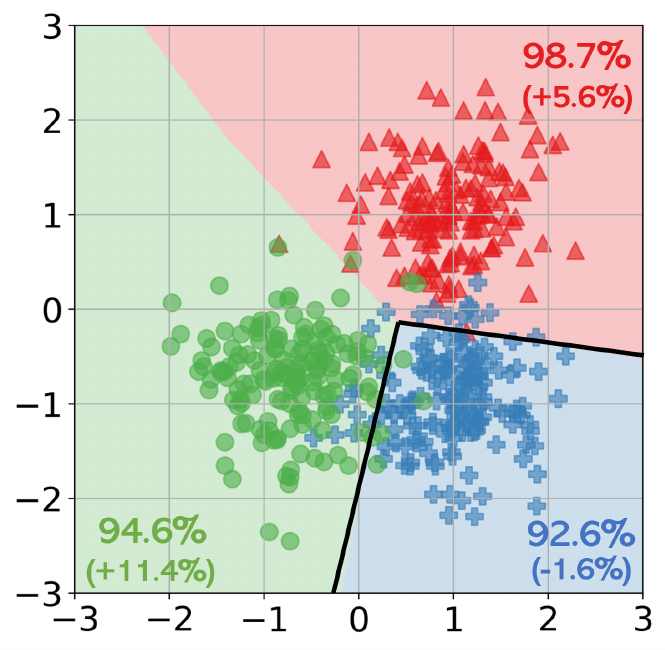}
    \vspace{-1.5em}
    \caption*{(c) UNSC}
  \end{minipage}
  \vspace{-0.5em}
  \caption{A toy example to illustrate the effectiveness of UNSC. (a) presents decision boundaries of the original model trained on four overlapping Gaussian distributions, with class-specific accuracy shown. We unlearn the entire purple class and plot the results in (b) and (c) for baseline and UNSC. Our method preserves the decision boundaries between remaining samples by unlearning in the null space and further improves the model's accuracy by calibrating the decision space.}
  \label{fig: toy example}
\end{figure}

However, current unlearning methods find that the unlearning process significantly affects the remaining data, leading to \textit{over-unlearning} \citep{hu2023duty,fan2023salun}. The model loses part of its prediction performance on the remaining data after unlearning. This is because the unlearning process requires modifications to the weights of the original model. Though the intent was to eliminate the influence of the unlearning data, these modifications inadvertently result in partial forgetting of the remaining samples, ultimately leading to over-unlearning. This raises a compelling question: 

\textit{Q1: To maintain the model's performance, can we prevent over-unlearning, and if so, how to achieve it?}

In this paper, we propose machine \underline{U}nlearning via \underline{N}ull \underline{S}pace \underline{C}alibration (UNSC). To avoid over-unlearning, UNSC constrains the unlearning process within a null space tailored to the remaining samples. This ensures unlearning does not negatively impact the model's performance on the remaining samples. Figure \ref{fig: toy example}c visualizes this concept, where decision boundaries among remaining samples closely resemble those depicted in Figure \ref{fig: toy example}a. Going beyond, we further ask:

\textit{Q2: Is it possible to improve the model's performance after unlearning, and how?}

To answer this question, we find that targeted unlearning can mitigate class overlap and augment the model's accuracy. Therefore, calibrating the decision space might be a possible solution. Take a toy case shown in Figure \ref{fig: toy example} as an example. We train a model to classify a mixture of four Gaussian distributions. Figure \ref{fig: toy example}a depicts the decision boundaries. Due to the overlap of the purple class with other classes, the model struggles to distinguish the overlapping samples. Unlearning misclassified samples and calibrating decision space lead to more accurate decision boundaries shown in Figure \ref{fig: toy example}c. 

UNSC assigns pseudo-labels to unlearning samples based on their proximity to the remaining samples. This process reallocates the space of unlearning samples to the remaining classes, forming more separable clusters (as shown in Figure \ref{fig: umap_unlearn_part}). Contributions of this paper are summarized as follows:

\begin{itemize}
    \item We tackle the challenging problem of over-unlearning. Going beyond, we investigate the possibility of boosting the model's performance after unlearning.
    
    \item We propose UNSC, a novel machine unlearning method that confines the unlearning process to a specific null space and assigns pseudo-labels to the unlearning samples. Combined, UNSC is not only free from over-unlearning but can also boost the model's performance.

    \item We provide theoretical justifications for the proposed method and empirically validate it with extensive experiments. The results show UNSC can provide equivalent and even better performance than retraining. 
\end{itemize}

\section{Related Works}
\subsection{Machine Unlearning}
Current unlearning algorithms have two branches: \textit{exact unlearning} and \textit{approximate unlearning} \citep{xu2023machine}. Exact unlearning ensures that distributions of model weights in an unlearned model are indistinguishable from those in a model retrained from scratch. \cite{cao2015towards} explore this concept within the realm of statistical query learning, \cite{ginart2019making} propose an efficient data elimination algorithm for $k$-means clustering. However, these methods can not be generalized to deep learning. To address this limitation, SISA \citep{bourtoule2021machine} divides the training data into several segments and trains sub-models on each segment. It only needs to retrain the affected segment when unlearning requests arise. Nevertheless, the computational overhead of the exact unlearning method remains substantial.

Unlike exact unlearning, approximate unlearning is bounded by an approximate guarantee. Works in this stream estimate the influence of unlearning samples and modify the model weights to achieve unlearning. For example, \cite{graves2021amnesiac} made the estimation through the gradient, \cite{guo2019certified} developed their method on the influence function \citep{koh2017understanding}, \cite{golatkar2020eternal} use the fisher information matrix in their design. \cite{lin2023erm} associate the unlearning samples with feature maps. While our work belongs to this category, it does not require the estimation of the influence of unlearning samples.

Works most closely related to ours are those by \citep{Chen_2023_CVPR,li2023subspace}. Both \cite{Chen_2023_CVPR} and our approach shift decision boundaries. The distinct advantage of our work lies in the design of a sophisticated null space, which maintains the decision boundaries of the remaining classes, leading to better performance. \cite{li2023subspace} also explore subspace-based unlearning; however, their approach does not capitalize on unlearning samples to guide the shifting of decision boundaries, an essential aspect of our method.

\subsection{Subspace Learning}
DNNs are usually over-parameterized as they have more parameters than input samples. However, the intrinsic dimension is much smaller \citep{allen2019convergence}. The weight updates happen in a much smaller subspace than the original parameter space. \cite{li2018measuring} find that optimizing in a reduced subspace reaches 90\% performance of regular SGD training. \cite{gur2018gradient} observe that after a short training period, the weight updates converge to a tiny subspace spanned by several top eigenvectors of the Hessian matrix. 

Learning in subspace has many applications. \cite{li2022low} optimize DNNs in 40-dimensional space and achieve comparable performance to regular training over thousands or even millions of parameters. In the context of meta-learning, \cite{lee2018gradient} employs a learned subspace within each layer's activation space where task-specific learners perform gradient descent. This subspace is tailored to be sensitive to task-specific requirements, facilitating more effective learning across different tasks. \cite{farajtabar2020orthogonal},\cite{GPM}, and \cite{wang2021training} applied subspace in continual learning to reduce the issue of catastrophic forgetting.  

\section{Preliminaries and Notation}
We use bold lowercase to denote vectors, e.g., $\mathbf{x}_i$, and italics in uppercase for space or set, e.g., $\mathcal{S}$. The training dataset is represented as $\mathcal{D}=\left\{\mathbf{x}_i,y_i\right\}_{i=1}^N \subseteq \mathcal{X}\times\mathcal{Y}$, where $\mathcal{X}\subseteq\mathbb{R}^d$ is the input space and $\mathcal{Y}=\left\{1,..., K\right\}$ is the label space. We further split $\mathcal{D}$ into unlearning set $\mathcal{D}_u$ and remaining set $\mathcal{D}_r$. We consider unlearning a neural network $f$ with $L$ layers. The input feature of input $\mathbf{x}_i$ at layer $l$ is denoted as $\mathbf{r}_l^i$. We denote $f(\cdot, \theta_o)$ the original model trained on $\mathcal{D}$ with parameters $\theta_o=\left\{\mathbf{w}_o^1,...,\mathbf{w}_o^L\right\}$, and $f(\cdot, \theta_r)$ is trained on $\mathcal{D}_r$. 

A machine unlearning algorithm takes the original model $f(\cdot, \theta_o)$ and the unlearning samples $\mathcal{D}_u$ as input and outputs an unlearned model $f(\cdot, \theta_u)$, in which the trace of the unlearning samples is removed. It can be formally defined as:
\begin{definition}[Machine unlearning \citep{cao2015towards}] Given a model $f(\cdot, \theta_o)=\mathcal{A}(\mathcal{D})$ trained on dataset $\mathcal{D}$ with some training algorithm $\mathcal{A}$, denote $\mathcal{D}_u \subseteq \mathcal{D}$ the set of samples that we want to remove from the training dataset $\mathcal{D}$ as well as from the trained model $\mathcal{A}(\mathcal{D})$. An unlearning process $\mathcal{U}$ produces a new set of weights $\theta_u$ that performs as though it had never seen the unlearning dataset $\mathcal{D}_u$.
\begin{equation}
    \theta_u \gets \mathcal{U}(f(\cdot, \theta_o), \mathcal{D}_u).
\end{equation}
\end{definition}

A key design of UNSC is a specific null space in which we perform unlearning. It ensures that unlearning will not affect the classification of the remaining samples. A null space is defined as:
\begin{definition}[Null space] Let $\mathbf{A}$ be an $m$ by $n$ matrix, the null space of $A$, denoted by $null(\mathbf{A})$, is the set of all vectors $\eta \in \mathbb{R}^n$ that satisfy $\mathbf{A}\eta=\mathbf{0}$.
\begin{equation}
    null(\mathbf{A})=\left\{\eta\in \mathbb{R}^n|\mathbf{A}\eta=\mathbf{0}\right\}.
\end{equation}
\end{definition}

\section{Proposed Method}
Our proposed method consists of two key parts: 1) unlearning in the null space to avoid over-unlearning and 2) calibrating decision space to improve the model's performance. 

\subsection{Find the Null Space for Unlearning} 
An ideal unlearning algorithm should erase the trace about the unlearning samples from the model without degrading the model’s performance on the remaining samples, \textit{i.e.}, without over-unlearning. To achieve this, UNSC constrains the unlearning process within a null space tailored to the remaining samples. The underlying principle is that different models with weights in the same null space would yield identical predictions on the samples associated with that null space (theoretical justification is provided in Section \ref{sec: analysis null space}). 

\vspace{0.3em}
\noindent\textbf{Gradients lie in the subspace spanned by inputs.} Considering a single-layer neural network performing a binary classification task. The loss function is defined as:
\begin{equation}
    L_i(\mathbf{w})=-y_i\log\left(\sigma(\mathbf{w}^T\mathbf{x}_i)\right)-(1-y_i)\log\left(1-\sigma(\mathbf{w}^T\mathbf{x}_i)\right),
\end{equation}
in which $\sigma(z)=\frac{1}{1+e^{-z}}$ is the sigmoid function. According to the chain rule, we have:
\begin{equation}
    \nabla_\mathbf{w} L_i=\left(\frac{1-y_i}{1-\sigma(\mathbf{w}^T\mathbf{x}_i)}-\frac{y_i}{\sigma(\mathbf{w}^T\mathbf{x}_i)}\right)\frac{\partial \sigma(\mathbf{w}^T\mathbf{x}_i)}{\partial\mathbf{w}}.
\end{equation}
Substitute the last term with the derivative of the sigmoid function, we get:
\begin{equation} \label{eq:g_x_linear}
    \nabla_\mathbf{w} L_i=\left(\sigma(\mathbf{w}^T\mathbf{x}_i)-y_i\right)\mathbf{x}_i=e_i\mathbf{x}_i,
\end{equation}
$e_i$ is the prediction error. Eq. \ref{eq:g_x_linear} shows the gradient lies in the space $span\left\{\mathbf{x}_1, \mathbf{x}_2,..., \mathbf{x}_N\right\}$. The same relation can be deduced for the convolutional layer by reformulating the convolution into matrix multiplication \citep{liu2018frequency}. 

\vspace{0.3em}
\noindent\textbf{Identify the layer-wise subspace for each class.} Based on the relationship between gradient and inputs, we first identify the subspace of each class. Then, for each unlearning sample, we find the corresponding null space to perform unlearning. This incurs negligible computation overhead. For each class, we only need a small batch of samples ($256$ in our experiment) to find the null space. In addition, we only need to find the subspace once because we treat the null space as fixed during unlearning. Our experiment results justify the hypothesis. One may consider iterative updating the null space during unlearning. We leave this as future work.

To find the subspace of class $k$, we sample a batch of inputs $\mathcal{B}_k=\left\{(x_i,y_i)|y_i=k\right\}_{i=1}^B$ from class $k$ at random and feed these samples through the original model $f(\cdot;\theta_o)$. At the $l$-th layer, we have a collection of input features as $\mathbf{R}_l^k=[\mathbf{r}_{l,1}, \mathbf{r}_{l,2},..., \mathbf{r}_{l,B}]$ with each column corresponds to an input. For the convolutional layer, we need to reformulate the input feature by treating each patch as a column. The columns of $\mathbf{R}_l^c$ are usually larger than the rows, and each column represents a vector sampled from some unknown subspace. We decompose it using singular value decomposition (SVD) to find the subspace as:
\begin{equation}
    \mathbf{U}_l^k\mathbf{\Sigma}_l^k\mathbf{V}_l^{kT}=\text{SVD}(\mathbf{R}_l^k).
\end{equation}

\begin{algorithm}[htb]
    \caption{Find layer-wise subspace of class $k$}
    \label{alg: subspce}
    \textbf{Input}: Batch input $\left\{x_i,y_i|y_i=k\right\}_{i=1}^B$; network $f(\cdot;\theta_o)$; threshold $\left\{\epsilon_l\right\}_{l=1}^L$.\\
    \textbf{Output}: Layer-wise subspace of class $k$.
    \begin{algorithmic}[1]
        \State Feed forward $\left\{x_i\right\}_{i=1}^B$ through $f(\cdot;\theta_o)$.
        \For{$l\in [L]$}
            \State Collect the input feature $\mathbf{R}_l^k$.
            \State Perform SVD: $\mathbf{R}_l = \mathbf{U}_l^k\mathbf{\Sigma}_l^k\mathbf{V}_l^{kT}$.
        \EndFor
        \State \Return $\mathcal{S}^k=\left\{\mathbf{U}_l^k\right\}_{l=1}^L$.
    \end{algorithmic}
\end{algorithm}

The elements in $\mathbf{U}_l^k$ are the subspace basis for class $k$ at layer $l$. The gradient induced by samples from class $k$ lies in this subspace. $\mathbf{\Sigma}_l^{k}$ is a diagonal matrix that stores the eigenvalues in descending order. We repeat this process of all layers in $f(\cdot;\theta_o)$ to identify subspaces of each layer,
\begin{equation}
    \mathcal{S}^k=\left\{\mathbf{U}_l^k\right\}_{l=1}^L.
\end{equation}
Algorithm \ref{alg: subspce} formalizes this process. Note that the inputs do not need to be from the same class. However, in our design, we find the layer-wise subspace $\mathcal{S}^k$ of each class for efficient unlearning. The unlearning samples could come from different classes. For each unlearning sample, we need to find the corresponding null space to perform unlearning. By identifying class-wise null space, we can efficiently identify each unlearning sample's null space. We articulate this in the following part.

\vspace{0.3em}
\noindent\textbf{Unlearning in the null space} After getting the layer-wise subspaces of each class, we are ready to find the null space for unlearning. For an unlearning sample $(x_i,y_i=c)\in \mathcal{D}_u$, we perform unlearning in the null space of the rest classes $\Bar{c} = \left\{i\in [K]|i\ne c\right\}$. Denote the concatenation of the rest subspaces as $\mathcal{S}^{\Bar{c}}_l=\cup_{i \in \Bar{c}}\mathbf{U}^i_l$ at layer-$l$. We merge these subspaces as:
\begin{equation} \label{eq: merge subspace}
\mathbf{U}_l^{\Bar{c}}\mathbf{\Sigma}_l^{\Bar{c}}\mathbf{V}_l^{\Bar{c}T}=\text{SVD}\left(\mathcal{S}_l^{\Bar{c}}\right).
\end{equation}
The basis of the merged subspace comes from the element of $\mathbf{U}_l^{\Bar{c}}$. $\mathcal{S}_l^{\Bar{c}}$ has few dominant eigenvalues and a large number of very small ones. The eigenvectors corresponding to those small eigenvalues are usually irrelevant. Thus, we can approximate the subspace by the span of top-$k$ eigenvectors. 
\begin{equation} \label{eq: k-rank approxi}
\left\|\left(\mathcal{S}_l^{\Bar{c}}\right)_k\right\|_F^2\geq\epsilon_l\left\|\mathcal{S}_l^{\Bar{c}}\right\|_F^2,
\end{equation}
where $\epsilon_l$ is the approximation threshold (we find $\epsilon_l>0.97$ is good enough in our experiment), $\|\mathbf{A}\|_F$ is Frobenius norm of matrix $\mathbf{A}$. The top-$k$ eigenvectors span the approximate subspace: 
\begin{equation}\label{eq: subspace span}
    \hat{\mathbf{S}}_l^{\Bar{c}} = span\left\{u_{l,1}^{\Bar{c}},u_{l,2}^{\Bar{c}},...,u_{l,k}^{\Bar{c}}\right\},
\end{equation}
where $u_{l,i}^{\Bar{c}}$ is the $i$-th element of $\mathbf{U}_l^{\Bar{c}}$. Gradient lies in $\hat{\mathbf{S}}_l^{\Bar{c}}$ will alter the model's response to samples from $\Bar{c}$, which may lead to over-unlearning. As such, we project the gradient to a space perpendicular to $\hat{\mathbf{S}}_l^{\Bar{c}}$, \textit{i.e.}, the null space of the rest samples. The projection matrix is defined as:
\begin{equation} \label{eq: proj mat}
    \mathbf{P}_l^{\Bar{c}} = \mathbf{I}-\mathbf{\hat{S}}_l^{\Bar{c}}(\mathbf{\hat{S}}_l^{\Bar{c})T}.
\end{equation}
At layer-$l$, the updating rule is modified as:
\begin{equation} \label{eq: psgd}
    \mathbf{w}_{t+1}^l=\mathbf{w}_{t}^l-\eta \mathbf{P}_l^{\Bar{c}} \nabla_{\mathbf{w}^l}\mathcal{L}^c,
\end{equation}
where $\mathcal{L}^c$ is the classification loss on unlearning sample $(x_i,y_i=c)$. Algorithm \ref{alg: subspce2} illustrates the pseudo-code.

\subsection{Calibrate Decision Space} \label{sec: pseudo labeling} 
To calibrate the decision space, we borrow the idea from poisoning attacks, which aim to decrease the victim model's performance by feeding it poisoned samples. There are several poisoning attacks, one of which is the label-flipping attack \citep{rosenfeld2020certified}. In this attack, the input feature $\mathbf{x}_i$ remains unchanged, and the label is carefully set as the most dissimilar class. But our target is different, \textit{i.e.}, we aim to boost the model's utility on the remaining samples. As such, we label an unlearning sample as the most similar class in the rest of the classes. 

However, comparing samples-wise similarity could be rather inefficient and even impossible when the remaining samples are not accessible. As such, we base our method on the prediction of the original model. For an unlearning sample of class $(\mathbf{x}_i,y_i=c)$, we find the pseudo-labels from the remaining classes set $\Bar{c}$ as:
\begin{align}
    \tilde{y} = \arg\max_{y\in \Bar{c}}f(\mathbf{x}_i;\theta_o).
\end{align}
We use the top-$1$ prediction as the pseudo-label for an unlearning sample if it gives the wrong prediction. Otherwise, we use the second-highest score as the pseudo-label.

\begin{algorithm}[!tb]
    \caption{Unlearning via null space calibration}
    \label{alg: subspce2}
    \textbf{Input}: Training data $\mathcal{D}$; trained model $f(\cdot;\theta_o)$.\\
    \textbf{Output}: Unlearned model $f(\cdot;\theta_u)$
    \begin{algorithmic}[1] 
        \State \# \textit{Get the subspace of each class}
        \For{$k \in [K]$}
            \State Identify the subspace of class $k$ by Algorithm \ref{alg: subspce} with $\left\{x_i,y_i|y_i=k\right\}_{i=1}^B$, $f(\cdot;\theta_o)$, and $\left\{\epsilon_l\right\}_{l=1}^L$ as input.
        \EndFor
        \State \# \textit{Calculate the projection matrix}
        \For{$c \in [K]$}
            \For{$l \in [L]$}
                \State Merge subspaces of rest classes using Eq. \ref{eq: merge subspace}.
                \State Perform $k$-rank approximation by Eq. \ref{eq: k-rank approxi}
                \State Find the subspace as Eq. \ref{eq: subspace span}.
                \State Calculate layer-wise projection matrix as Eq. \ref{eq: proj mat}.
            \EndFor
        \EndFor
        \State \# \textit{Unlearning in the null space}
        \For{$(x_i, y_i)\in \mathcal{D}_u$}
            \State $\tilde{y}_i$ $\gets$ $\arg\max_{y\in \Bar{c}}f(\mathbf{x}_i;\theta_o)$ \Comment{Pseudo-labeling $x_i$}
            \State Calculate gradients on ($x_i,\tilde{y}_i$).
            \State Update the original model in the null space as Eq. \ref{eq: psgd}.
        \EndFor
    \end{algorithmic}
\end{algorithm}

To elucidate our design, one might consider this in another way: we expect the unlearned model to behave similarly to the retrained model. The unlearned model should give the same or similar predictions as the retrained model for the unlearning samples. The retrained model, trained using the remaining samples, has never been exposed to the unlearning samples. Consequently, the retrained model will classify these unlearning samples into categories represented within the remaining samples. As such, the unlearned model should predict a given unlearning sample belongs to the same class as the remaining samples most similar to it.

\section{Theoretical Analysis}
\subsection{Null Space Prevents Over-unlearning}
\label{sec: analysis null space}
\begin{restatable}{theorem}{tns}\label{thm: null space}
    For a trained model $f(\cdot; \theta_o)$, unlearning the target samples from $\mathcal{D}_u$ in the null space tailored to the remaining samples from $\mathcal{D}_r$ ensures the unlearned model $f(\cdot; \theta_u)$ has approximately the same performance as $f(\cdot; \theta_o)$, \textit{i.e.},
    \begin{equation}
        \mathcal{L}_{r}(\theta_u) \approx \mathcal{L}_{r}(\theta_o),
    \end{equation}
    where $\mathcal{L}_{r}(\theta) := \frac{1}{|\mathcal{D}_r|}\sum_{\mathcal{D}_r}\ell(f(\mathbf{x}_i;\theta),y_i)$ is the average loss on the remaining samples.
\end{restatable}
\vspace{-1em}
\begin{proof}
We expand $\mathcal{L}_{r}(\theta_u)$ around $\mathcal{L}_{r}(\theta_o)$ as:
    \begin{align*}
        \mathcal{L}_{r}(\theta_u) &= \mathcal{L}_{r}(\theta_o) + (\theta_u-\theta_o)^T\nabla_{\theta}\mathcal{L}_{r}(\theta_o) + \mathcal{O}(\left\|\theta_u-\theta_o\right\|^2) \\ 
        &=\mathcal{L}_{r}(\theta_o) + \Delta \theta^T\nabla_{\theta}\mathcal{L}_{r}(\theta_o) + \mathcal{O}(\left\|\Delta \theta\right\|^2).       
    \end{align*}
We decompose $\Delta \theta^T\nabla_{\theta}\mathcal{L}_{r}(\theta_o)$ layer-by-layer as:
\begin{equation}
    \Delta \theta^T\nabla_{\theta}\mathcal{L}_{r}(\theta_o) = \frac{1}{|\mathcal{D}_r|}\sum_{l\in[L]} \sum_{\mathcal{D}_r}(\Delta \mathbf{w}^{l})^T\nabla_{\mathbf{w}^l}\ell(f(\mathbf{x}_i;\theta_o),y_i).
\end{equation}
Note $\theta=\left\{\mathbf{w}^1,...,\mathbf{w}^L\right\}$ and $\Delta \mathbf{w}^{l}=\mathbf{w}_u^1-\mathbf{w}_o^1$. At the $l$-th layer, we have $\mathbf{r}_i^{l+1} = \sigma(\mathbf{w}^l\mathbf{r}_i^{l})$, where $\sigma(\cdot)$ is activation function, $\mathbf{r}^l_i$ is the activation at layer $l$ with respect to input $\mathbf{x}_i$.  We rewrite the derivative term by the chain rule:
\begin{align*}
&\nabla_{\mathbf{w}^l}\ell(f(\mathbf{x}_i;\theta_0), y_i) =\\      &\nabla_f\ell(f(\mathbf{x}_i;\theta_0),y_i)\nabla_{w^{l+1}}f(\mathbf{x}_i;\theta_0)\text{diag}\left\{\sigma'(\mathbf{w}^l\mathbf{r}_i^l)\right\}\mathbf{r}_i^l. 
\end{align*}
Note that we confine the learning in a null space tailored to the reaming samples from $\mathcal{D}_r$. Thus at the $l$-th layer, we have the following by definition:
\vspace{-0.5em}
\begin{equation}\label{eq: inner product}
    \left<\Delta \mathbf{w}^l, \mathbf{r}^l_i\right>=0,\forall \mathbf{x}_i\in \mathcal{D}_r.
\end{equation}
Thus, we immediately have $\Delta \theta^T\nabla_{\theta}\mathcal{L}_{r}(\theta_o)=0$. During unlearning, $\Delta \theta$ should be relatively small. As such, we have:
\begin{equation}
    \mathcal{L}_{r}(\theta_u) \approx \mathcal{L}_{r}(\theta_o).
\end{equation}
\vspace{-2em}
\end{proof}
Theorem \ref{thm: null space} asserts that confining unlearning in the null space ensures the unlearned model $f(\cdot, \theta_u)$ performs comparably to the original model $f(\cdot, \theta_o)$. This substantiates that UNSC can effectively mitigate the risk of over-unlearning.

\subsection{Pseudo-labeling Benefits Unlearning} \label{sec: analysis pseudo labeling} 
Assuming the unlearning sample $\mathbf{x}_i$ is generated i.i.d. from the domain $\left<\mathcal{D}_u, h_o\right>$, with the ground-truth labeling function $h_o$. In our proposed method, we assign pseudo-labels to the unlearning samples, this equivalent to replacing $h_o$ with a new labeling function $h_u$. Consequently, the domain of unlearning samples becomes $\left<\mathcal{D}_u, h_u\right>$ during learning. The primary objective of unlearning is to minimize the model's retention of information about the unlearning samples. We use the empirical risk on unlearning samples as a proxy measurement of forgetting in our analysis. The risk of $f(\cdot;\theta_u)$ over $\left<\mathcal{D}_u, h_u\right>$ is calculated as:
\begin{equation}
    \epsilon(\theta_u,h_u) = \mathbb{E}_{\mathbf{x}_i\in \mathcal{D}_u}[\mathbf{1}_{f(\mathbf{x}_i;\theta_u)\neq h_u(\mathbf{x}_i)}].
\end{equation}
We use the parallel notion $\epsilon_u(\theta_u,h_o)$ to denote the risk over the original domain $\left<\mathcal{D}_u, h_o\right>$, which can be expressed as:
\begin{align}
    \epsilon(\theta_u,h_o) & = \epsilon(\theta_u,h_o) + \epsilon(\theta_u,h_u) - \epsilon(\theta_u,h_u) \nonumber \\ 
    & = \epsilon(\theta_u,h_u) + \mathbb{E}_{\mathbf{x}_i\in \mathcal{D}_u}[\mathbf{1}_{f(\mathbf{x}_i;\theta_u)\neq h_o(\mathbf{x}_i)}] \nonumber\\ 
    &- \mathbb{E}_{\mathbf{x}_i\in \mathcal{D}_u} [\mathbf{1}_{f(\mathbf{x}_i;\theta_u)\neq h_u(\mathbf{x}_i)}] \nonumber \\ 
    & \geq \epsilon(\theta_u,h_u) - \mathbb{E}_{\mathbf{x}_i\in \mathcal{D}_u}[\mathbf{1}_{h_o(\mathbf{x}_i) = h_u(\mathbf{x}_i)}] \label{eq: th2}.
\end{align}

The forgetting level $\epsilon(\theta_u,h_0)$ depends partly on the second term on the right-hand side. It measures the expectation that the pseudo-labeling function disagrees with the ground-truth labeling function for the unlearning samples. To facilitate forgetting, $h_u$ should assign each unlearning sample a label different from $h_o(\mathbf{x}_i)$ to minimize $\mathbb{E}_{\mathbf{x}_i\in \mathcal{D}_u}[\mathbf{1}_{h_o(\mathbf{x}_i) = h_u(\mathbf{x}_i)}]$. The pseudo-label assigned by UNSC differs from the ground-truth label $h_o(\mathbf{x}_i)$. Thus, it minimizes the second term and increases $\epsilon(\theta_u,h_o)$, leading to better forgetting.

\section{Experiments}
\label{sec: exp}
\subsection{Protocol and Evaluation Metrics}
\noindent\textbf{Datasets and models.} We compare UNSC with existing methods on the standard FashionMNIST \citep{xiao2017fashion}, CIFAR-10, CIFAR-100 \citep{cifar}, and SVHN \citep{svhn} benchmarks. We use AlexNet \citep{alexnet} for FashionMNIST, VGG-11 \citep{vgg} for SVHN, AllCNN \citep{allcnn} for CIFAR-10, and ResNet-18 \citep{resnet} for CIFAR-100.

\vspace{0.3em}
\noindent\textbf{Evaluation metrics.} We evaluate the unlearning process from two aspects, \textit{i.e.}, utility and privacy guarantee. (1) \textit{Utility guarantee} is evaluated by accuracy on the remaining testing data $Acc_{\mathcal{D}_{rt}}$ and accuracy on the unlearning testing data $Acc_{\mathcal{D}_{ut}}$. An ideal unlearning process should sustain $Acc_{\mathcal{D}_{rt}}$ and reduce $Acc_{\mathcal{D}_{rt}}$ towards zero \citep{Chen_2023_CVPR}. (2) \textit{Privacy guarantee} is assessed through the membership inference attack (MIA) \citep{shokri2017membership}. Following \citep{fan2023salun}, we apply the confidence-based MIA predictor to the unlearned model on the unlearning samples. We define the MIA accuracy as the True Negative Rate $Acc_{MIA}=TN/|\mathcal{D}_u|$, \textit{i.e.}, the rate of unlearning samples that are identified as not being in the training set of the unlearned model. A higher $Acc_{MIA}$ indicates a more effective unlearning. All results are averaged over three trials by default. 

\vspace{0.3em}
\noindent\textbf{Implementation details.}
The original training set is divided into $90\%$ for training and $10\%$ for validation. We train the original and retrained models for $200$ epochs and stop the training based on validation accuracy. The patience is set to $30$. We use SGD optimizer in all experiments, starting with a learning rate of $0.1$, and reducing it by $0.2$ at epochs $60$, $120$, and $160$. In the unlearning stage, we determine the best learning rate and number of epochs for different datasets and networks. Table \ref{tab: exp detail} in Appendix \ref{app: exp detail} details the protocol. 

\subsection{Comparison with the State-of-the-Art}
We compare UNSC with various methods, assessing both utility and privacy guarantees. These include: (1) \textit{Retrain}, (2) \textit{Boundary unlearning (BU) }\citep{Chen_2023_CVPR}, (3) \textit{Random labels (RL)} \citep{hayase2020selective}, (4) \textit{Gradient ascent (GA)} \citep{golatkar2020eternal}, (5) \textit{Fisher Forgetting (Fisher)} \citep{golatkar2020eternal}, and (6) \textit{SalUn} \citep{fan2023salun}. Following \citep{Chen_2023_CVPR}, we randomly select one class (or ten classes on CIFAR-100) and unlearn all the samples. 

\vspace{0.3em}
\noindent\textbf{Utility guarantee.} Table \ref{tab: effec} displays the accuracy results of models obtained by different unlearning methods. Observing the results reveals that: (1) \textit{Comparison methods struggle with over-unlearning}: $Acc_{\mathcal{D}_{ut}}$ decreases at the cost of $Acc_{\mathcal{D}_{rt}}$, leading to over-unlearning. For instance, SalUn reduces $Acc_{\mathcal{D}_{ut}}$ to $3.56\%$ with a $16.7\%$ reduction in $Acc_{\mathcal{D}_{rt}}$ on the SVHN dataset. This is more sever on the CIFAR-100 dataset, as comparison methods degrade $Acc_{\mathcal{D}_{rt}}$ by more than $30\%$; (2) \textit{UNSC excels in avoiding over-unlearning}: UNSC consistently maintains $Acc_{\mathcal{D}_{ut}}$ below 2\% without compromising $Acc_{\mathcal{D}_{rt}}$. This underscores UNSC's ability to avoid over-unlearning; (3) \textit{UNSC exhibits superior performance compared to the original model}: On all datasets, models derived from UNSC get higher $Acc_{\mathcal{D}_{rt}}$ than the original models. This improvement can be attributed to the calibration of decision space, which makes the remaining samples more separable. 

\begin{figure}[t]
    \centering
    \begin{subfigure}[b]{0.23\textwidth}
        \includegraphics[width=\textwidth]{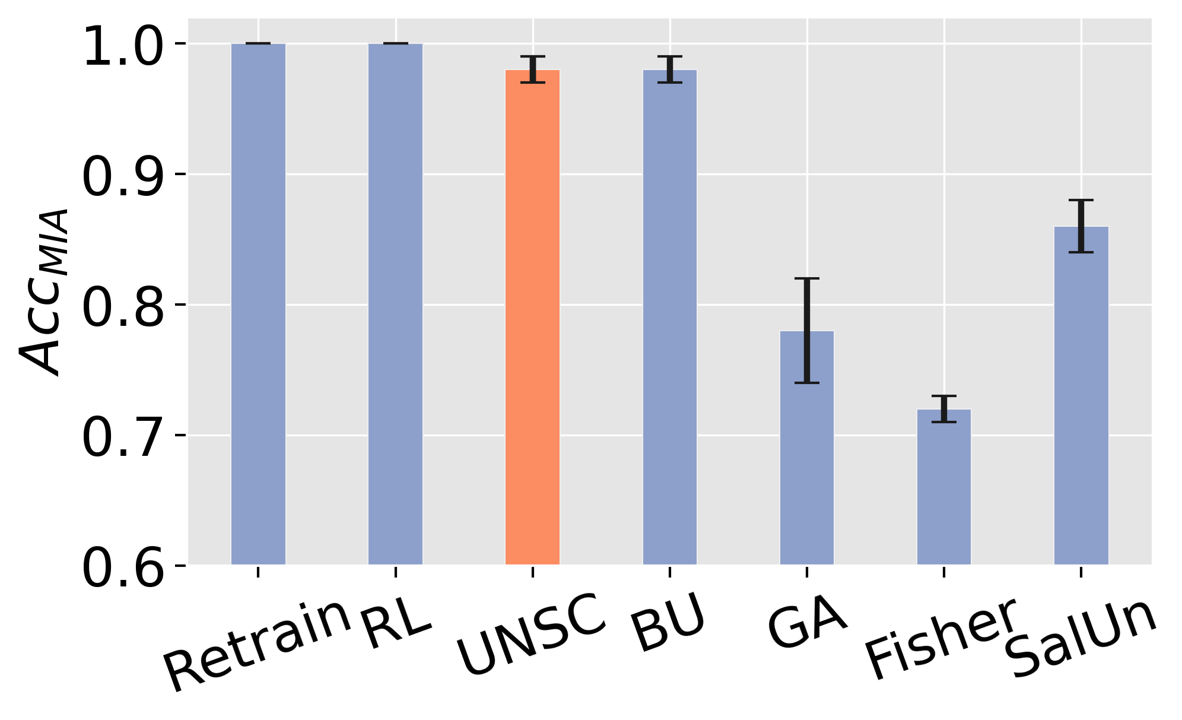}
        \vspace{-1.5em}
        \caption{On CIFAR-10.}
        \label{fig: mia cifar10}
    \end{subfigure}
    \hfill
    \begin{subfigure}[b]{0.23\textwidth}
        \includegraphics[width=\textwidth]{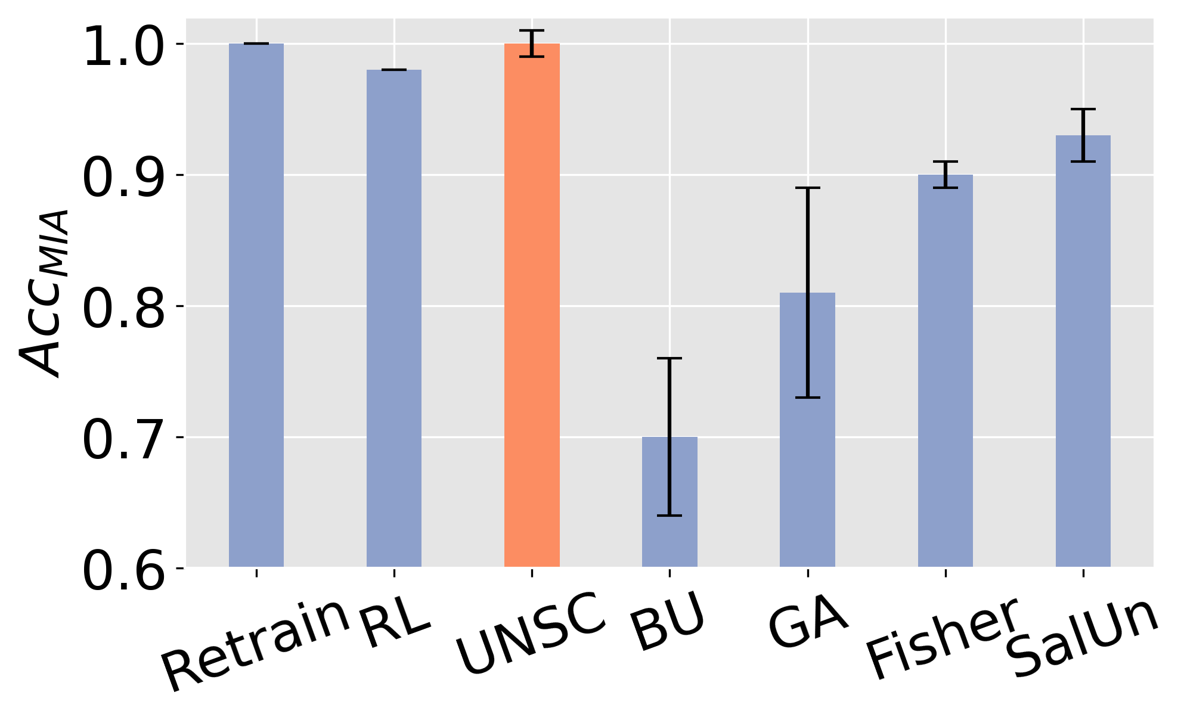}
        \vspace{-1.5em}
        \caption{On SVHN.}
        \label{fig: mia svhn}
    \end{subfigure}
    \vspace{-0.8em}
    \caption{Comparing the MIA accuracy of different unlearning methods. A higher $Acc_{MIA}$ indicates more effective unlearning.}
    \vspace{-0.5em}
    \label{fig: mia}
\end{figure}

\vspace{0.3em}
\noindent\textbf{Privacy guarantee.} Figure \ref{fig: mia} depicts MIA results. By randomly selecting one class to unlearn, we get various unlearned models through different methods. We then assess these models using MIA. The retrained model, having no exposure to unlearning samples, obtains an attack accuracy of $1.0$. On CIFAR-10, Fisher and GA exhibit an accuracy below $0.8$, pointing to inadequate unlearning. SalUn achieves a higher accuracy but suffers more severe over-unlearning than Fisher, as evident by $Acc_{\mathcal{D}_{rt}}$ in Table \ref{tab: effec}. The attack accuracy of BU and UNSC is around $0.98$, signifying success unlearning. However, \textit{unlike BU, UNSC is free of over-unlearning.} For SVHN, the attack accuracy of BU and GA falls below $0.8$, suggesting unsuccessful unlearning. Conversely, UNSC gets an attack accuracy of $1.0$. This validates the effectiveness and robustness of UNSC.

\begin{table*}[!thp]
    \centering
    \caption{Comparing the utility guarantee among UNSC and SOTA methods: For FashionMNIST, SVHN, and CIFAR-10, we randomly unlearn one class, and for CIFAR-100, we unlearn ten classes. The results of the retrained model are provided as a reference.}
    \label{tab: effec}
    \setlength{\tabcolsep}{4pt}
    \small
    \begin{tabular}{c|cc|cc|cc|cc}
    \toprule
   \multirow{2}{*}{Approach} & \multicolumn{2}{c}{FashionMNIST} & \multicolumn{2}{c}{SVHN} & \multicolumn{2}{c}{CIFAR-10} & \multicolumn{2}{c}{CIFAR-100}\\
    & $Acc_{\mathcal{D}_{rt}}(\uparrow)$ & $Acc_{\mathcal{D}_{ut}}(\downarrow)$ & $Acc_{\mathcal{D}_{rt}}(\uparrow)$ & $Acc_{\mathcal{D}_{ut}}(\downarrow)$ & $Acc_{\mathcal{D}_{rt}}(\uparrow)$ & $Acc_{\mathcal{D}_{ut}}(\downarrow)$ &
    $Acc_{\mathcal{D}_{rt}}(\uparrow)$ & $Acc_{\mathcal{D}_{ut}}(\downarrow)$\\
    \midrule
    \midrule
    Original & $93.58 _{\pm 0.28}$ & $80.97 _{\pm 1.44}$ & $95.52_{\pm0.12}$ & $91.30_{\pm0.30}$ & $92.10 _{\pm 0.29} $ & $79.93 _{\pm 1.72}$ & $75.36_{\pm 0.34}$ & $72.07 _{\pm 1.18}$  \\
    RL & $91.50 _{\pm 0.20}$ & $3.37 _{\pm 1.44} $ & $75.87_{\pm 8.67} $ & $1.17_{\pm 0.16} $ & $82.88 _{\pm 2.62} $ & $3.30 _{\pm 0.83} $ & $24.58 _{\pm 2.01} $ & $1.13 _{\pm 0.61}$\\
    BU & $72.71 _{\pm 0.44} $ & $2.97 _{\pm 1.28} $ & $80.99 _{\pm 4.01} $ & $4.03 _{\pm 1.87} $ & $87.24 _{\pm 2.43} $ & $1.47 _{\pm 0.17} $ & $36.81 _{\pm 3.05} $ & $12.13 _{\pm 2.40} $ \\
    GA & $80.94 _{\pm 3.87} $ & $6.40 _{\pm 2.90} $ & $ 75.11 _{\pm 6.76} $& $3.10 _{\pm 3.90} $& $88.53 _{\pm 1.11} $ & $ 6.80 _{\pm 1.93} $ & $36.55_{\pm12.85} $ & $12.87_{\pm 7.61}$\\
    Fisher & $95.11 _{\pm 0.17}$ & $1.50 _{\pm 1.27}$& $95.72 _{\pm 0.30} $ & $2.53 _{\pm 0.91} $ & $92.54 _{\pm 0.79} $ & $0.07 _{\pm 0.05} $ & $49.23 _{\pm 2.02} $ & $0.00_{\pm0.00}$ \\
    SalUn &$91.11 _{\pm 0.32} $ & $1.80 _{\pm 1.44} $ & $78.81 _{\pm 4.85} $ & $3.56 _{\pm 0.56} $ & $88.88 _{\pm 0.19}$ & $6.17 _{\pm 0.83}$ & $41.85 _{\pm 3.56}$ & $5.97 _{\pm 1.39}$ \\
    \rowcolor{lightgray} UNSC & $\textbf{95.23}_{\pm\textbf{0.27}}$ & $0.67 _{\pm 0.17}$ & $\textbf{95.74} _{\pm \textbf{0.17}}$ & $0.00_{\pm0.00}$ & $92.77 _{\pm 0.08}$ & $0.70 _{\pm 0.08} $ & $\textbf{76.16} _{\pm \textbf{0.35}} $ & $1.80 _{\pm 0.16} $ \\
    \midrule
    Retrain & $95.19 _{\pm 0.20}$ & $0.00 _{\pm 0.00} $ & $95.56_{\pm0.23}$ & $0.00 _{\pm0.00}$ & $\textbf{93.22} _{\pm \textbf{0.18}}$ & $0.00_{\pm 0.00} $ & $75.77_{\pm 0.22} $ & $0.00_{\pm0.00}$ \\
    \bottomrule
    \end{tabular} %
\end{table*}

\subsection{Ablation Studies}
\label{exp: pseudo-labeling}
\noindent\textbf{Pseudo-labeling strategy.} To demonstrate the efficacy of the proposed pseudo-labeling strategy, we conduct a thorough examination of different datasets. We present results obtained on CIFAR-10 in Figure \ref{fig: pseudo labeling} and provide results on other datasets in Appendix \ref{app: additional results}. We unlearn all samples of class $3$ and assign pseudo-labels as described in Section. \ref{sec: pseudo labeling}. Figure \ref{fig: pseudo labeling} plots distributions of pseudo-labels and labels predicted by the retrained model. The distribution of pseudo-labels aligns well with the predictions of the retrained model on unlearning samples. The consistency substantiates the effectiveness of the proposed pseudo-labeling strategy. 

\begin{figure}[t]
    \centering
    \begin{subfigure}[b]{0.23\textwidth}
        \scalebox{1.0}[0.85]{\includegraphics[width=\textwidth]{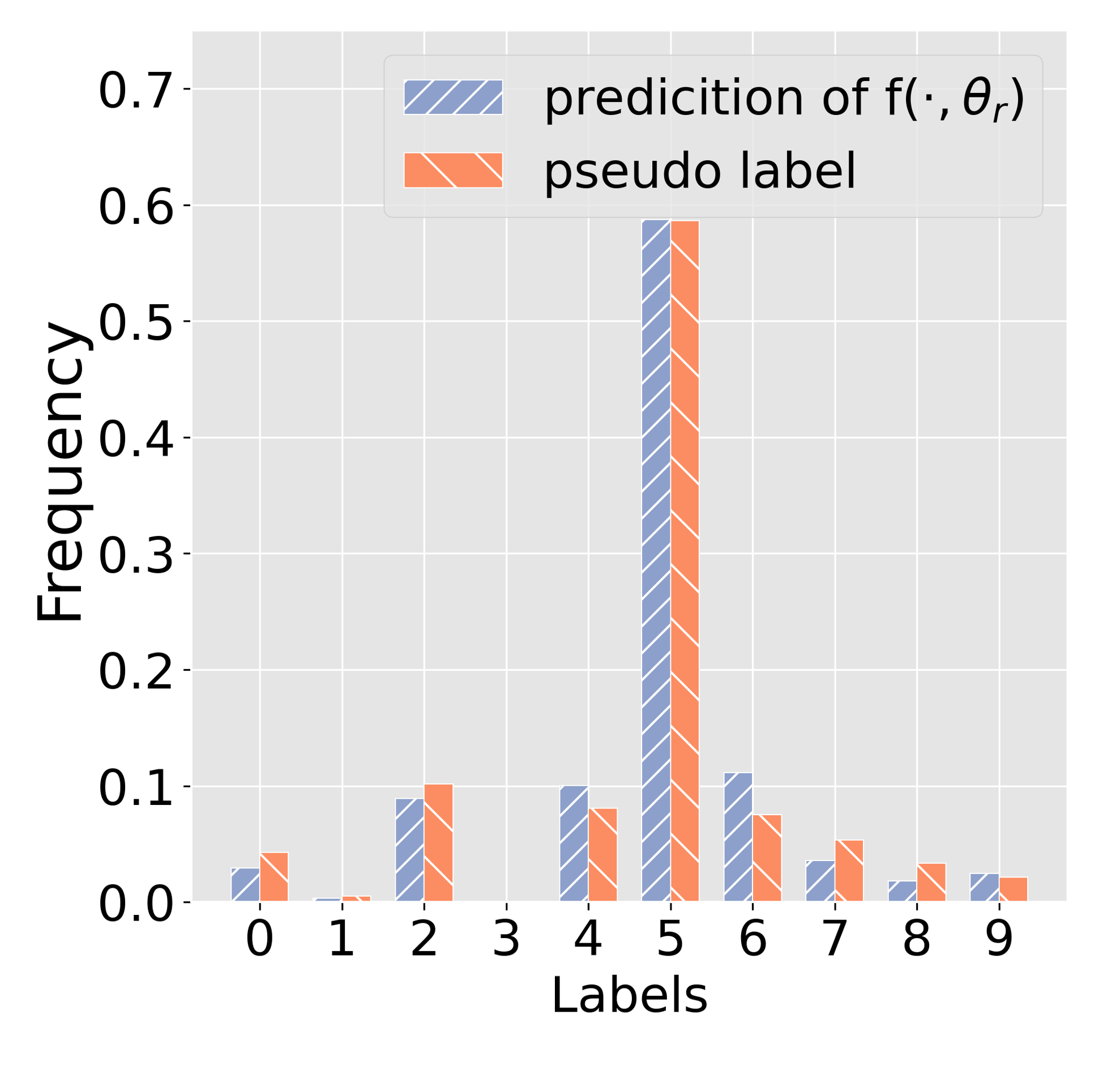}}
        \vspace{-1.7em}
        \caption{}
        \label{fig: pseudo labeling}
    \end{subfigure}
    \hfill
    \begin{subfigure}[b]{0.23\textwidth}
        \scalebox{1.0}[0.85]{\includegraphics[width=\textwidth]{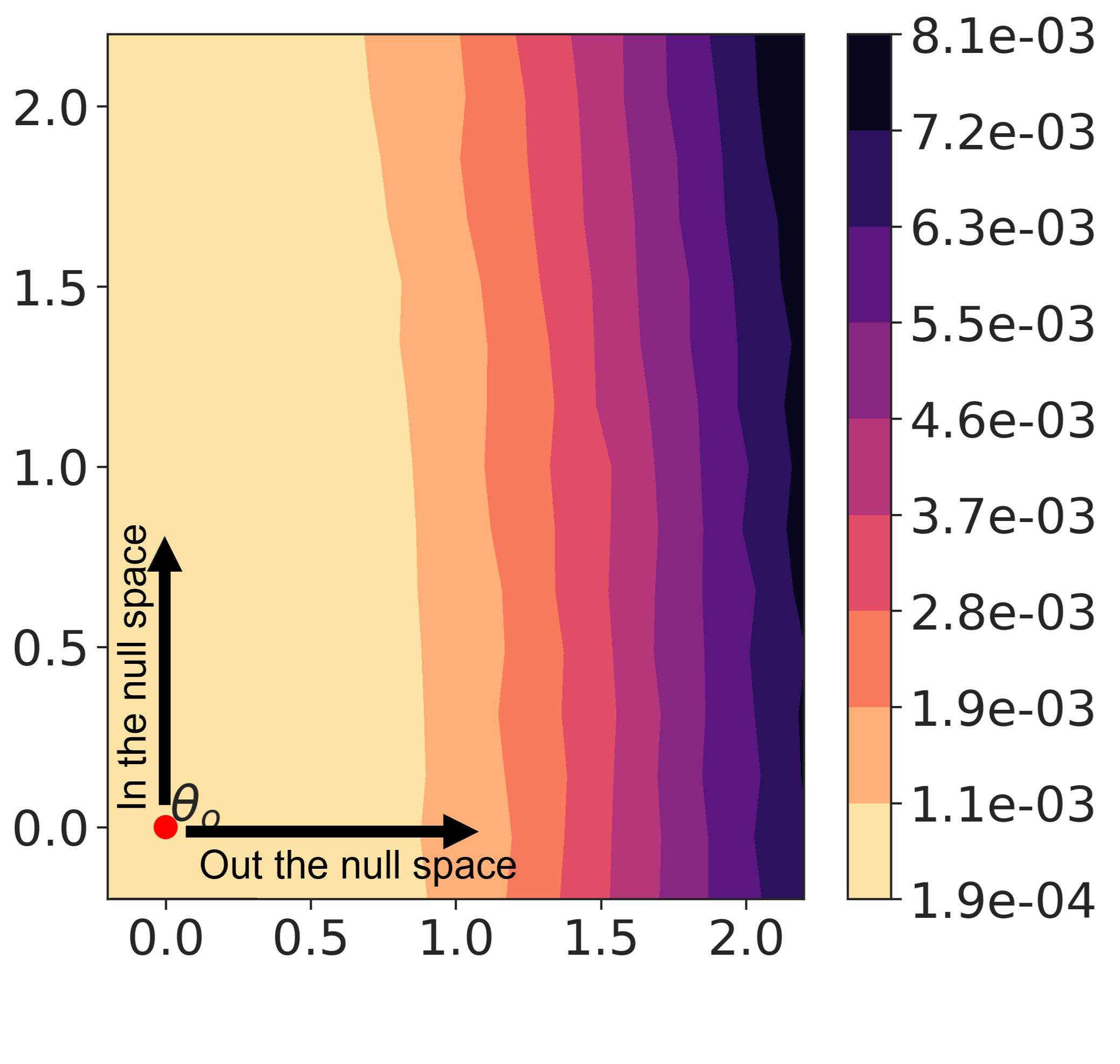}}
         \vspace{-1.7em}
        \caption{} 
            \label{fig: loss contour}
    \end{subfigure}
    \vspace{-0.5em}
    \caption{(a) Distributions of pseudo-labels and predictions of the retrained model on CIFAR-10. (b) Loss contour on CIFAR-10. We evaluate the model's loss on remaining test data.} 
    \vspace{-1.5em}
\end{figure}

\vspace{0.3em}
\noindent\textbf{Impact of null space.} Our first contribution is confining the unlearning process in a null space, which avoids over-unlearning. Here, we provide experimental validation. We randomly select one class as the unlearning class and identify the corresponding null space following the steps outlined in Algorithm \ref{alg: subspce}. We modify the model's weights by adding vectors composed of two components: one within the null space and one outside. We evaluate the modified model on the testing set of remaining samples and plot the loss contour in Figure \ref{fig: loss contour}, with the axes displaying the scaled coefficients. 

From Figure \ref{fig: loss contour}, we observe that modifying weights outside the null space increases the loss, while alterations within the null space leave the loss unchanged. This pattern validates our analysis: unlearning within the null space does not impact the model's performance on the remaining samples. Results on additional datasets are presented in Appendix \ref{app: additional results}.

\begin{figure}[t]
    \centering
    \begin{subfigure}[b]{0.2\textwidth}
        \scalebox{1}[0.9]{\includegraphics[width=\textwidth]{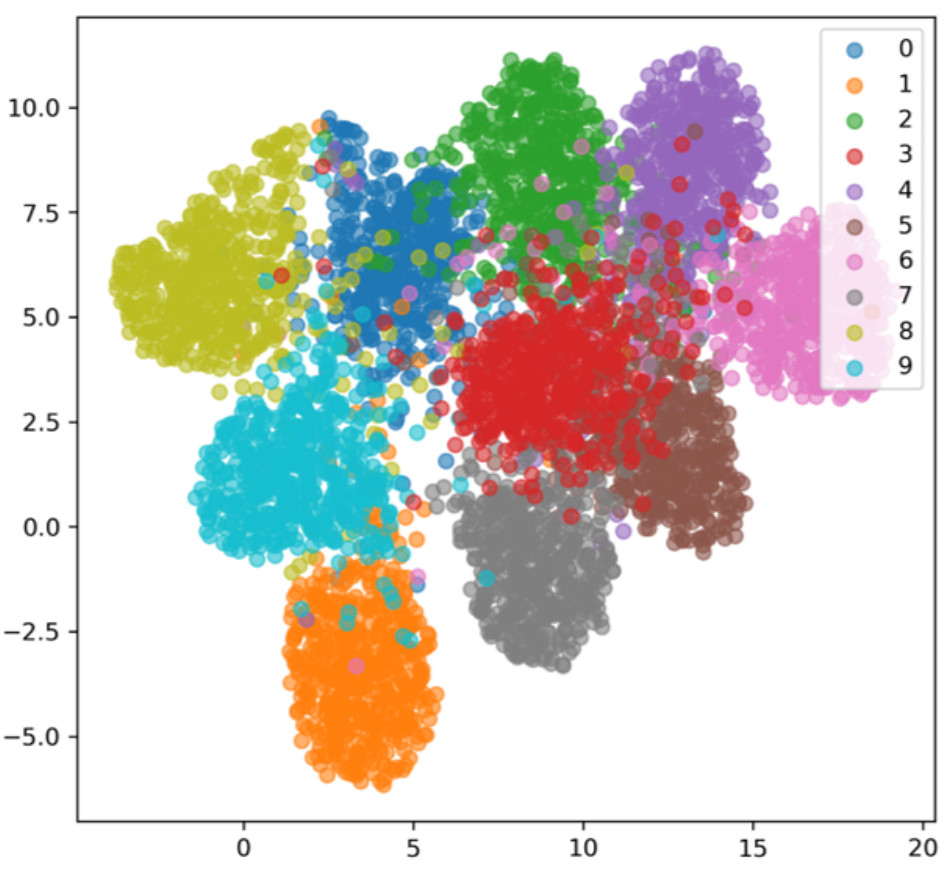}}
        \vspace{-1.5em}
        \caption{Original model (full)} 
        \label{fig: umap_ori_full}
    \end{subfigure}
    \hspace{0.8em}
    \begin{subfigure}[b]{0.2\textwidth}
        \scalebox{1}[0.9]
        {\includegraphics[width=\textwidth]{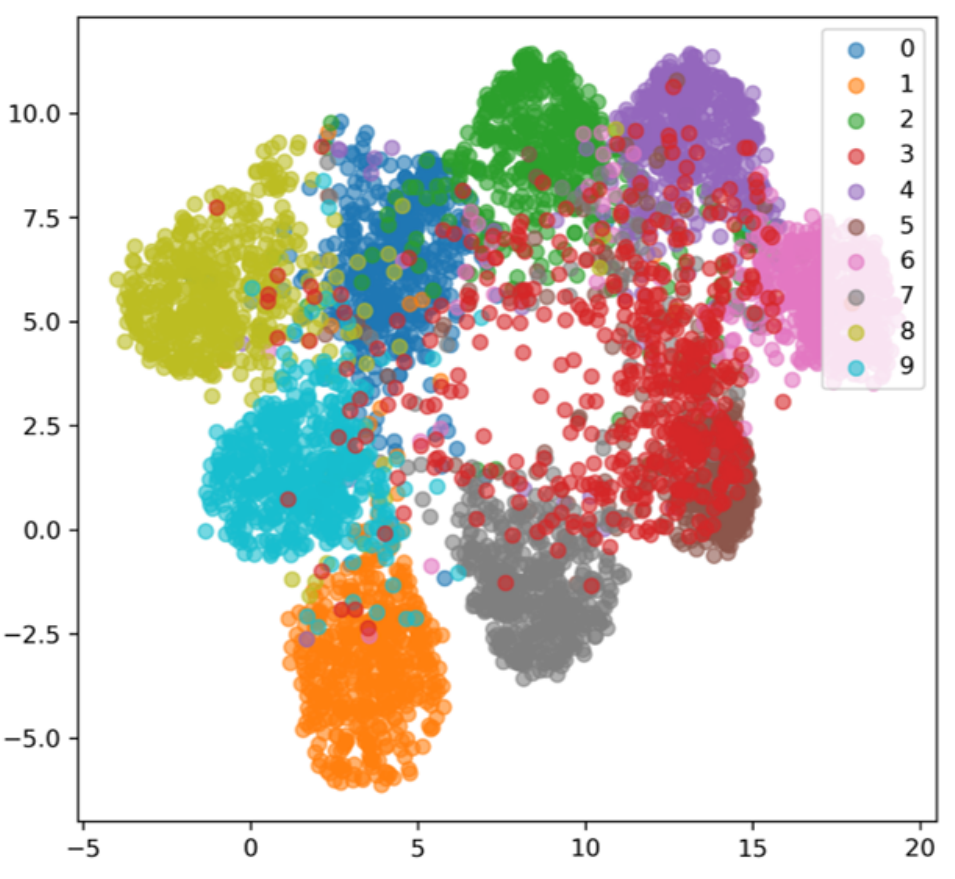}}
        \vspace{-1.5em}
        \caption{Unlearned model (full)} 
        \label{fig: umap_unlearn_full}
    \end{subfigure}
    \vspace{0.3em}
    \\
    \begin{subfigure}[b]{0.2\textwidth}
    \scalebox{1}[0.9]
        {\includegraphics[width=\textwidth]{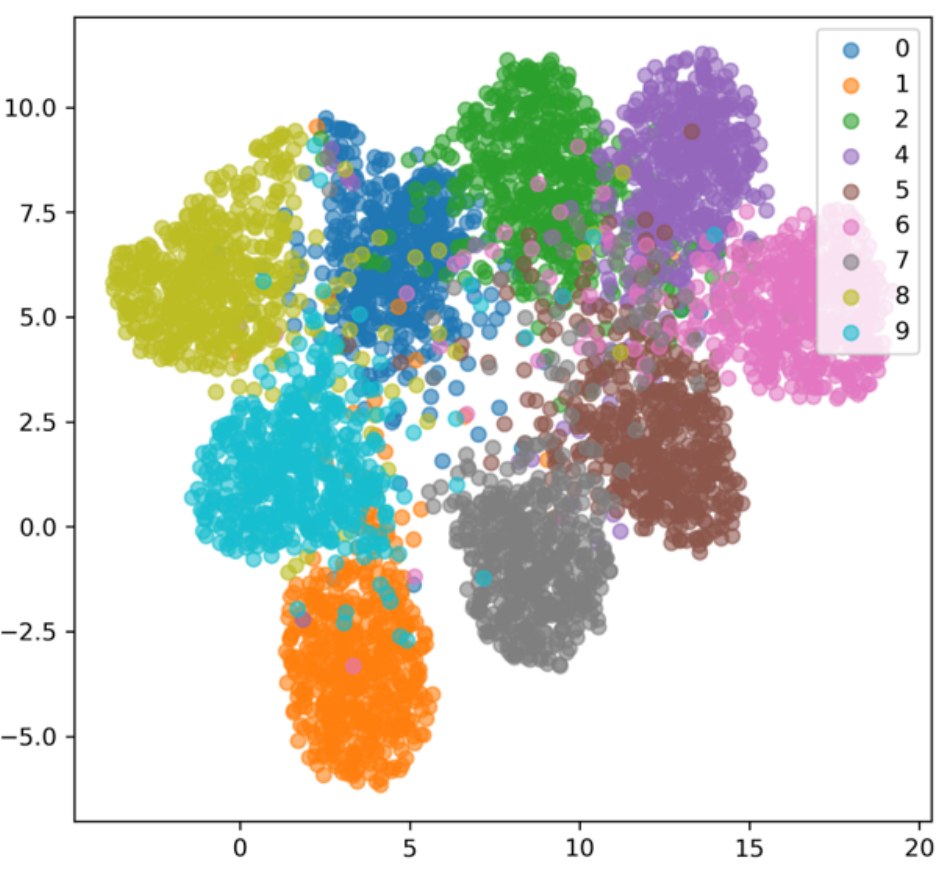}}
        \vspace{-1.5em}
        \caption{Original model (part)} 
        \label{fig: umap_ori_part}
    \end{subfigure}
    \hspace{0.8em}
    \begin{subfigure}[b]{0.2\textwidth}
    \scalebox{1}[0.9]
        {\includegraphics[width=\textwidth]{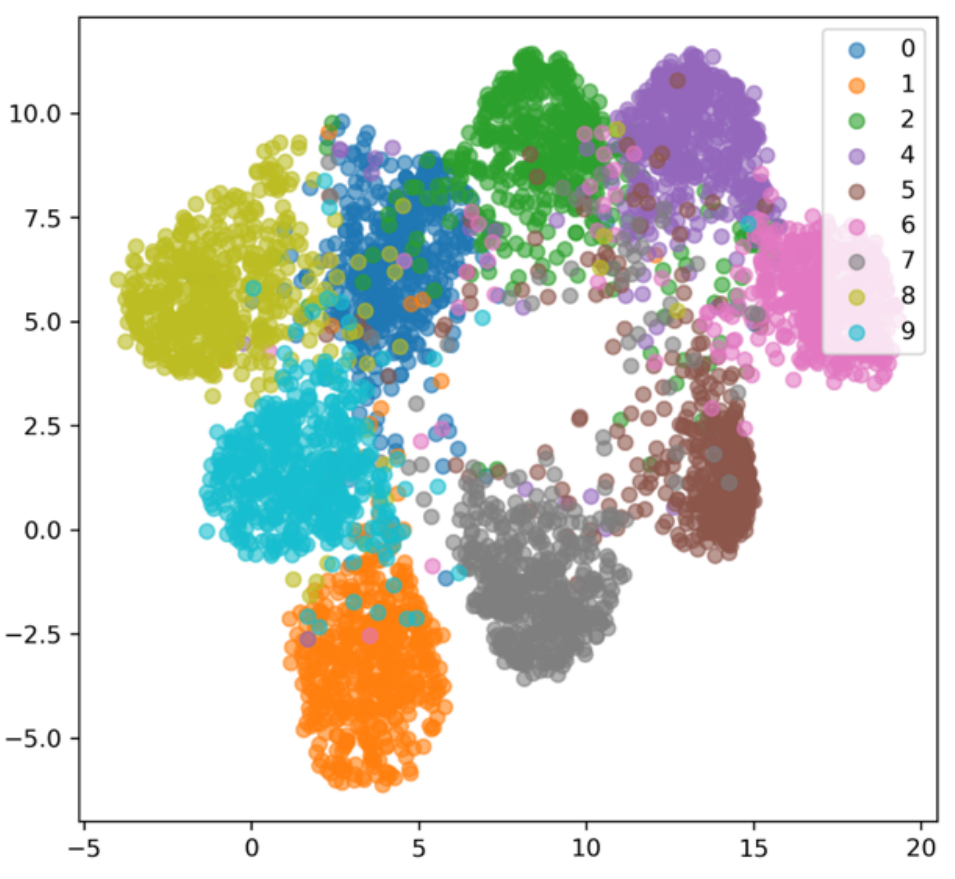}}
        \vspace{-1.5em}
        \caption{Unlearned model (part)} 
        \label{fig: umap_unlearn_part}
    \end{subfigure}
    \vspace{-0.5em}
    \caption{Visualization of latent space after unlearning all samples of class $3$ (red points) on CIFAR-10: (a) Original and (b) Unlearned model with all samples, (c) Original and (d) Unlearned models, both excluding the unlearned samples.}
    \vspace{-1.5em}
    \label{fig: umap}
\end{figure}

\vspace{0.3em}
\noindent\textbf{Decision space calibration.} Our second contribution is the calibration of the decision space, which improves the model's performance on the remaining samples. To illustrate this, we compare decision spaces of the original and unlearned model in Figure \ref{fig: umap}. We unlearn class $3$ and visualize the latent spaces using UMAP \citep{mcinnes2018umap}. Several observations can be made: (1) \textit{UNSC successfully unlearns class $3$ from the original model}. In Figure \ref{fig: umap_unlearn_full}, the cluster of class 3 disperses, with its samples overlapping with adjacent samples. A clear signal that the unlearned model forgets the unlearned samples; (2) \textit{Decision boundaries among the remaining samples remain unaffected after unlearning}. The shape and positions of clusters in Figure \ref{fig: umap_unlearn_part} closely resemble those in Figure \ref{fig: umap_ori_part}, underscoring the efficacy of UNSC in overcoming over-unlearning; (3) \textit{Better separability after unlearning}. UNSC redistributes the region previously dominated by class $3$ among neighboring classes, improving their separability as depicted in Figures \ref{fig: umap_ori_part} and \ref{fig: umap_unlearn_part}. 

\begin{figure*}
    \centering
     \begin{subfigure}[b]{0.23\textwidth}
        \includegraphics[width=\textwidth]{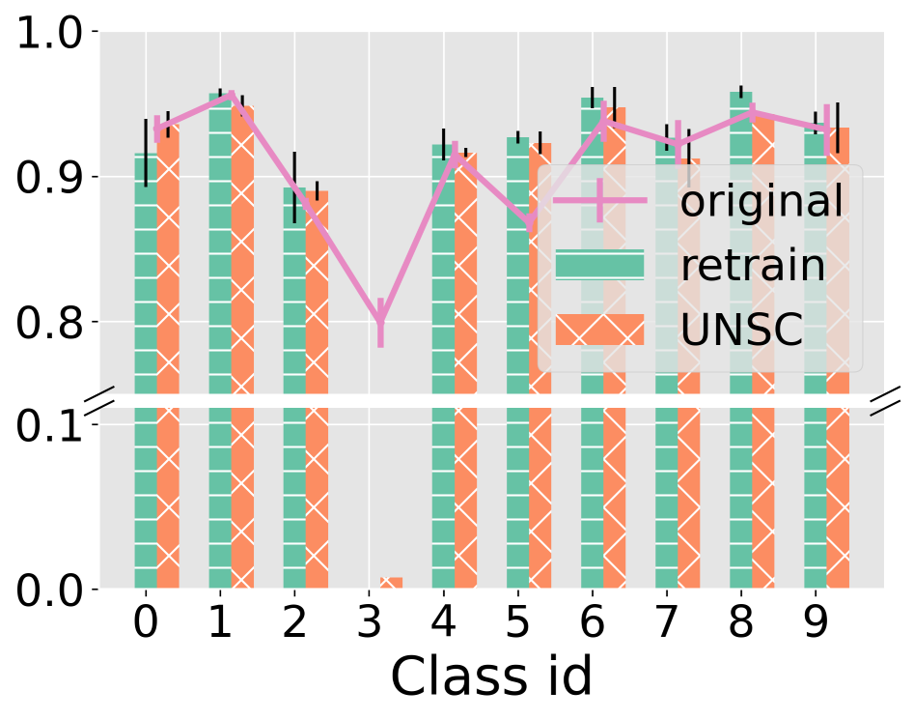}
        \vspace{-1.7em}
        \caption{Unlearn $1$ class} 
        \label{fig: cls-acc-CIFAR-10-1}
    \end{subfigure}
    \begin{subfigure}[b]{0.23\textwidth}
        \includegraphics[width=\textwidth]{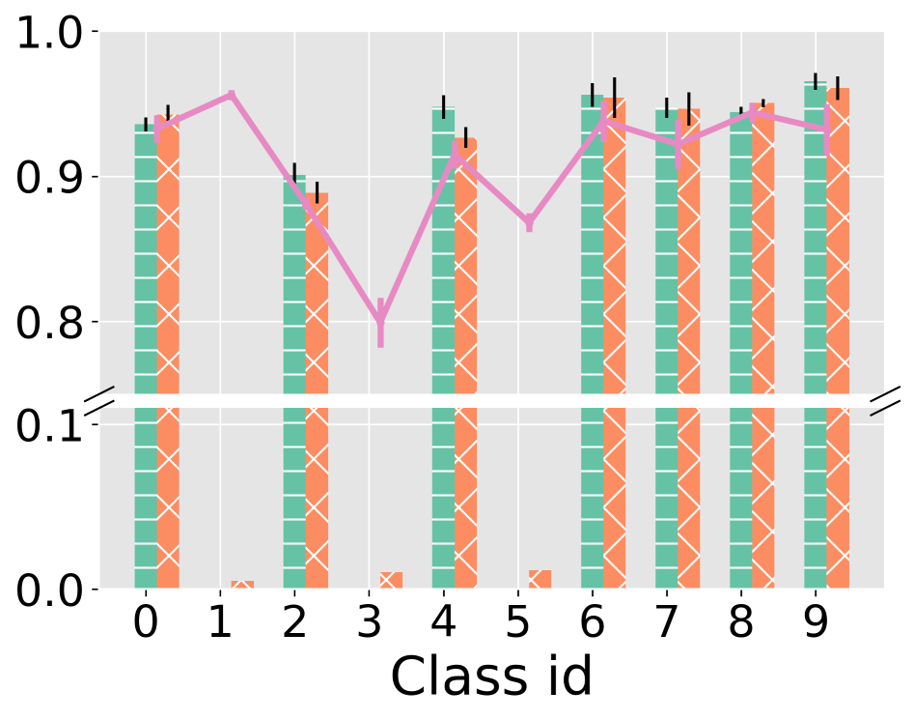}
        \vspace{-1.7em}
        \caption{Unlearn $3$ classes} 
        \label{fig: cls-acc-CIFAR-10-3}
    \end{subfigure}
    \begin{subfigure}[b]{0.23\textwidth}
        \includegraphics[width=\textwidth]{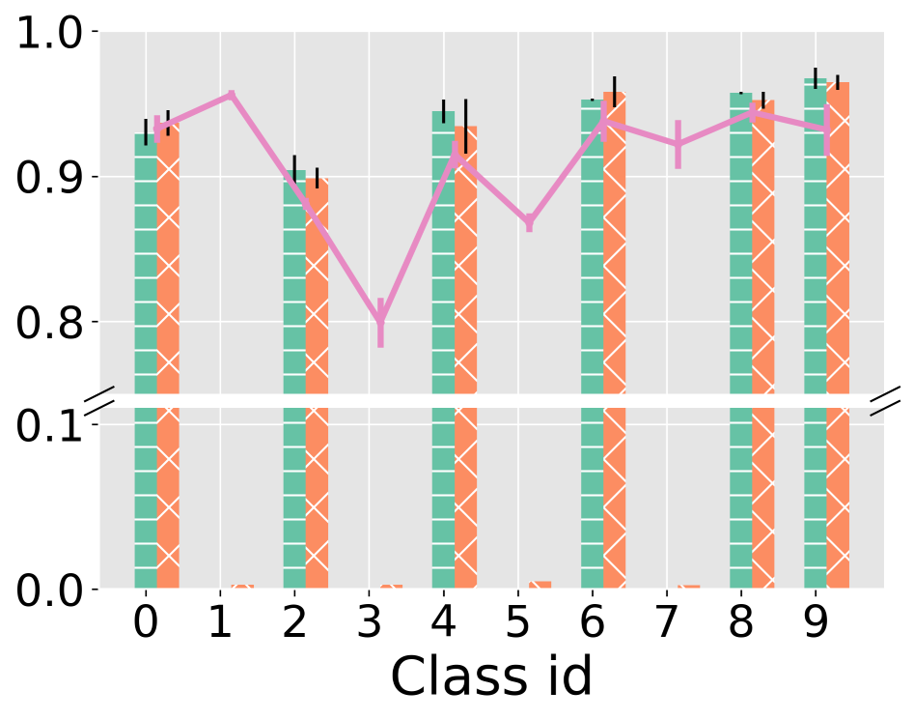}
        \vspace{-1.7em}
        \caption{Unlearn $4$ classes} 
        \label{fig: cls-acc-CIFAR-10-4}
    \end{subfigure}
    \begin{subfigure}[b]{0.23\textwidth}
        \includegraphics[width=\textwidth]{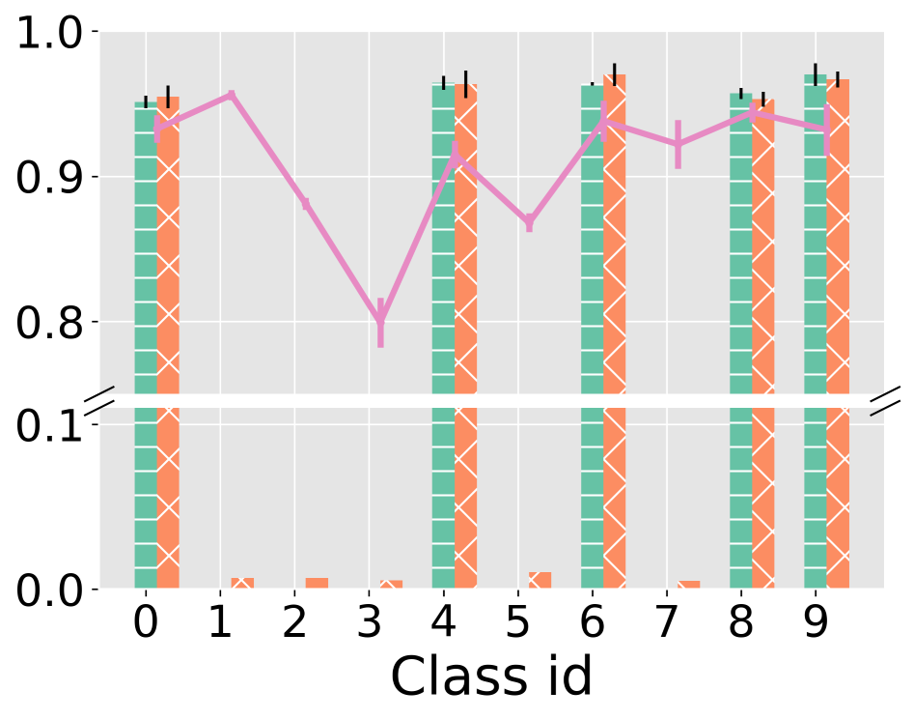}
        \vspace{-1.7em}
        \caption{Unlearn $5$ classes} 
        \label{fig: cls-acc-CIFAR-10-5}
    \end{subfigure}
    \caption{Comparison of class-wise accuracy on CIFAR-10. We compare the original model, the retrained model, and the model obtained by UNSC under different settings. In all experiments, the UNSC-obtained model demonstrates near-zero accuracy on the forgetting classes, achieving superior accuracy compared to the original model and comparable performance to the retrained model.}
    \label{fig: cls-acc-CIFAR-10}
\end{figure*}

\begin{table*}
    \centering
    \caption{Performance evaluation of UNSC on CIFAR-10. \#Classes denotes the number of unlearned classes.} 
    \footnotesize
    \resizebox{0.75\linewidth}{!}{
    \label{tab: multi-classes unlearn}
    \setlength{\tabcolsep}{4pt}
    \begin{tabular}{c|ccccccc}
         \toprule
         \multirow{2}{*}{\#Classes} & \multicolumn{2}{c}{Original} & \multicolumn{2}{c}{Retrain} & \multicolumn{2}{c}{UNSC} & Accuracy Gain\\
         & $Acc_{\mathcal{D}_{rt}}$ & $Acc_{\mathcal{D}_{ut}}$ & $Acc_{\mathcal{D}_{rt}}(\uparrow)$ & $Acc_{\mathcal{D}_{ut}}(\downarrow)$ & $Acc_{\mathcal{D}_{rt}}(\uparrow)$ & $Acc_{\mathcal{D}_{ut}}(\downarrow)$ & $\Delta Acc_{\mathcal{D}_{rt}}(\uparrow)$\\
         \midrule
         \midrule
         1 & $92.10 _{\pm 0.29} $ & $79.93 _{\pm 1.72} $ & $93.22 _{\pm 0.18} $ & $ 0.00 _{\pm 0.00} $ & $92.77 _{\pm 0.25} $ & $0.70 _{\pm 0.08} $ & $0.67_{\pm 0.42} $\\
         2 & $91.66 _{\pm 0.36} $ & $87.77_{\pm 0.89} $ & $93.18 _{\pm 0.15} $ & $0.00 _{\pm 0.00} $ & $92.73 _{\pm 0.25} $ & $0.60 _{\pm 0.33} $ & $1.07_{\pm0.35}$\\
         3 & $92.35 _{\pm 0.35} $ & $87.44 _{\pm 0.64} $ & $94.24 _{\pm 0.17} $ & $0.00 _{\pm 0.00} $ & $93.84 _{\pm 0.33} $ & $0.89 _{\pm 0.45} $ & $1.49_{\pm0.57}$\\
         4 &$92.38 _{\pm 0.36} $  & $88.63 _{\pm 0.77} $ & $94.28 _{\pm 0.36} $ & $0.00 _{\pm 0.00} $ & $94.08 _{\pm 0.21} $ & $0.31 _{\pm 0.23} $ & $1.70_{\pm 0.54} $\\
         5 &$93.23 _{\pm 0.34}$ & $88.53 _{\pm 0.55}$ & $96.12 _{\pm 0.22}$ & $0.00 _{\pm 0.00}$ & $96.15 _{\pm 0.13}$ & $0.68 _{\pm 0.52}$ & $2.92_{\pm0.42}$\\
         \bottomrule
    \end{tabular}
    }
\end{table*}

\vspace{0.5em}
\noindent\textbf{UNSC under different settings.} We randomly select $1$ to $5$ classes to unlearn. Table \ref{tab: multi-classes unlearn} compares the utility guarantee of the original model, the retrained model, and the model obtained by UNSC on the CIFAR-10 dataset. Figure \ref{fig: cls-acc-CIFAR-10} provides a detailed class-wise accuracy breakdown. Additional results on other datasets can be found in Appendix \ref{app: additional results}. 

Across different settings, the model obtained by UNSC gets near-zero accuracy on the unlearned classes. Meanwhile, the accuracy of the remaining classes is higher than that of the original model (measured by $\Delta Acc_{\mathcal{D}_{rt}}$). Furthermore, as the number of unlearned classes increases, so does the accuracy gain. This observation affirms (1) the effectiveness of null space in preventing over-unlearning and (2) the benefit of decision space calibration. As we progressively unlearn additional samples, we allocate more decision spaces to the remaining samples. This process results in a more distinguishable decision space, ultimately enhancing the model's classification performance on the remaining samples.

\vspace{0.3em}
\noindent\textbf{UNSC vs. Baselines.} To discern the contribution of each component in the proposed method, we contrast UNSC with three baselines: the original model, Retrain, and RL. Table \ref{tab: ablation fmnist} presents a detailed breakdown of the results on the FashionMNIST dataset, with indexed experiments for reference. (for results on other datasets, please refer to Appendix \ref{app: additional results}). 

\textit{Contribution of null space.} We conduct an ablation study by removing the constraint of null space and then compare the outcomes of experiments III and IV. In the absence of null space constraints, RL reduces $Acc_{\mathcal{D}_{ut}}$ to $0.90\%$, incurring a $7.28\%$ reduction in $Acc_{\mathcal{D}_{rt}}$. RL successfully completes the unlearning task but at the cost of excessive impact on the remaining samples. In contrast, when confining the unlearning process within a null space tailored to the remaining samples in experiment IV, we decrease $Acc_{\mathcal{D}_{ut}}$ to $1.57\%$ without affecting $Acc_{\mathcal{D}_{rt}}$. This comparison highlights the effectiveness of the null space in preventing over-unlearning.

\textit{Contribution of pseudo-labeling.} We further demonstrate the benefits introduced by pseudo-labeling. Compared to experiment IV, replacing RL with pseudo-labeling brings a $1.93\%$ improvement in $Acc_{\mathcal{D}_{rt}}$ in experiment V. Thanks to pseudo-labeling, the decision space of the unlearning samples can be selectively redistributed to the adjacent classes, forming a more distinguishable decision space. The combination of null space and pseudo-labeling contributes to an elevated $Acc_{\mathcal{D}_{rt}}$ of $95.23\%$, surpassing the original model by $1.61\%$ and even outperforming the retrained model.

\begin{table}[t]
    \centering
    \caption{Ablation results on FashionMNIST. }
    \label{tab: ablation fmnist}
    \setlength{\tabcolsep}{3pt}
    \resizebox{.48\textwidth}{!}{
    \begin{tabular}{c|ccccc|cc}
        \toprule
         Exp. ID & Original & Retrain & RL & \begin{tabular}[c]{@{}c@{}}Null\\ space\end{tabular} & \begin{tabular}[c]{@{}c@{}}Pesudo-\\labeling\end{tabular}  & $Acc_{\mathcal{D}_{rt}}$ & $Acc_{\mathcal{D}_{ut}}$\\
         \midrule
         \midrule
         I & $\checkmark$ & & & & &$93.58 _{\pm 0.28} $ & $80.97 _{\pm 1.44} $\\
         II & & $\checkmark$ & & & &$95.19 _{\pm 0.20} $ & $0.00_{\pm0.00}$\\
         III & & & $\checkmark$& & & $86.30_{\pm 0.85}$ & $0.90_{\pm 0.21}$\\
         \midrule
         IV & &  & $\checkmark$ & $\checkmark$ &  & $93.30 _{\pm 1.12}$ & $1.57 _{\pm 0.53}$\\
         V & &  & & $\checkmark$ & $\checkmark$& $95.23 _{\pm 0.27}$ & $0.67_{\pm 0.17}$\\
         \bottomrule
    \end{tabular}
    }
\end{table}
\vspace{-0.3em}

\section{Conclusion}
This paper introduces UNSC, an accurate unlearning algorithm that addresses the challenge of over-unlearning and further enhances the model's performance after unlearning. The innovation in UNSC is grounded in two pivotal design aspects: (1) Constraining the unlearning process within a designated null space to prevent over-unlearning and (2) Pseudo-labeling the unlearning samples to calibrate the decision space, thereby improving prediction accuracy. Our theoretical analysis provides robust support for this approach, and empirical results convincingly demonstrate the superior performance of UNSC over existing methods.

\appendix

\bibliographystyle{named}
\bibliography{main}

\clearpage
\onecolumn 
\begin{nolinenumbers}
\begin{center}
\textbf{\large Appendix}
\end{center}
\renewcommand{\thesection}{A} 
\section{Experiment Setups}
\label{app: exp detail}
We validate UNSC on several datasets using multiple networks. We split the original training set with 90\% as the training set and the remaining 10\% as the validation set. We monitor the model's accuracy on the validation set during training and set early stop the training with patience set as $30$. We train the original model $\theta_o$ and retrained model $\theta_r$, with SGD optimizer using an initial learning rate of $0.1$ and decrease it by 0.2 at milestones of epoch $60, 120$ and $160$, respectively. In the unlearning stage, we chose the best learning rate for different datasets\&networks. Table \ref{tab: exp detail} lists the training details.
\begin{table*}[h]
    \centering
    \caption{Experiment details}
    \label{tab: exp detail}
    \resizebox{0.85\linewidth}{!}{
    \begin{tabular}{c|ccccccccc}
    \toprule
   \multirow{2}{*}{Dataset}
    & \multirow{2}{*}{Network} & \multirow{2}{*}{Batch size} & \multirow{2}{*}{Optimizer} & \multicolumn{3}{c}{Training $\theta_o$ and $\theta_r$} & \multicolumn{3}{c}{Unlearning}\\
    & & & & lr & wd & epoch & lr & wd & epoch\\
    \midrule
    FashionMNIST & AlexNet & $512$ & SGD & $0.1$ & $0.0005$ & $200$ & $0.0005$ & $0$ & $15$ \\
    SVHN & VGG11 & $512$ & SGD & $0.1$ & $0.0005$ & $200$ & $0.03$ & $0$ & $15$ \\
    CIFAR-10 & AllCNN & $512$ & SGD & $0.1$ & $0.0005$ & $200$ & $0.04$ & $0$ & $25$\\
    CIFAR-100 & ResNet18 & $512$ & SGD & $0.1$ & $0.0005$ & $200$ & $0.04$ & $0$ & $25$\\
    \bottomrule
    \end{tabular}
    }
\end{table*}

\renewcommand{\thesection}{B}
\section{Additional Results}
\label{app: additional results} 
\subsection{UNSC under different settings}
For the FashionMNIST and SVHN datasets, we randomly select $1$ to $5$ classes to unlearn. Tables \ref{tab: multi-classes unlearn fmnist} and \ref{tab: multi-classes unlearn svhn} present the results, and Figures \ref{fig: cls-acc-fmnist} and \ref{fig: cls-acc-svhn} show the comparisons of class-wise accuracy. On the CIFAR-100 dataset, we randomly chose $10,20,30,40,50$ classes to unlearn. Table \ref{tab: multi-classes unlearn CIFAR-100} records the results on CIFAR-100, and Figure \ref{fig: cls-acc-CIFAR-100} plots the class-wise accuracy for $10$ class case. On all datasets, UNSC can successfully unlearn the target classes (as verified by the low $Acc_{\mathcal{D}_{ut}}$) and, in the meantime, get a higher performance than the original model on the remaining classes, \textit{i.e.}, higher $Acc_{\mathcal{D}_{rt}}$. 

\begin{figure*}[htp]
    \centering
     \begin{subfigure}[b]{0.24\textwidth}
        \includegraphics[width=\textwidth]{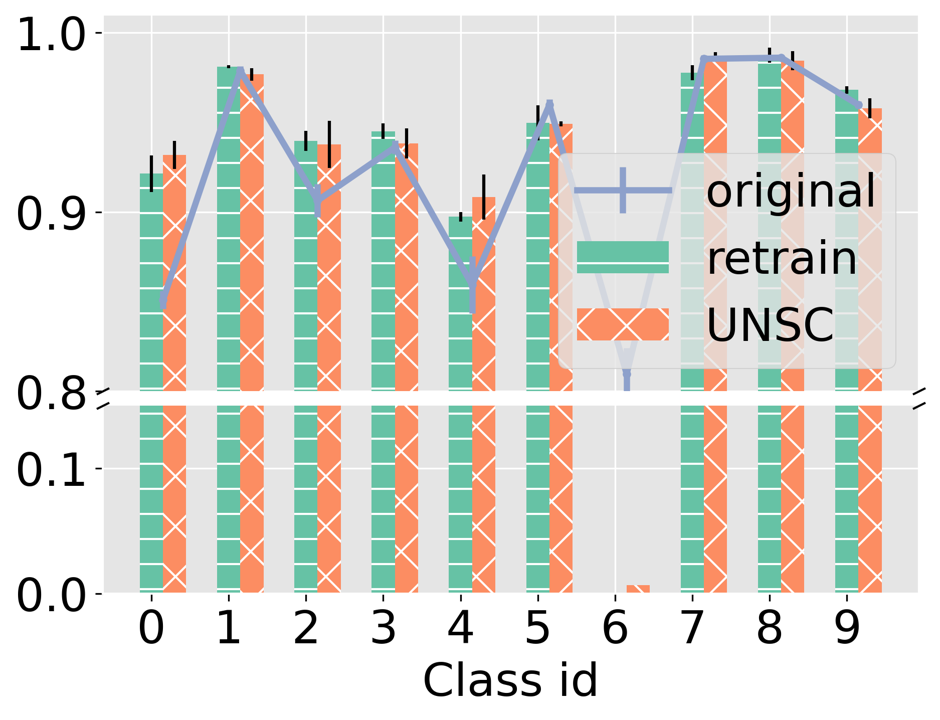}
        \caption{Unlearn $1$ class} 
        \label{fig: cls-acc-fmnist-1}
    \end{subfigure}
    \hfill
         \begin{subfigure}[b]{0.24\textwidth}
        \includegraphics[width=\textwidth]{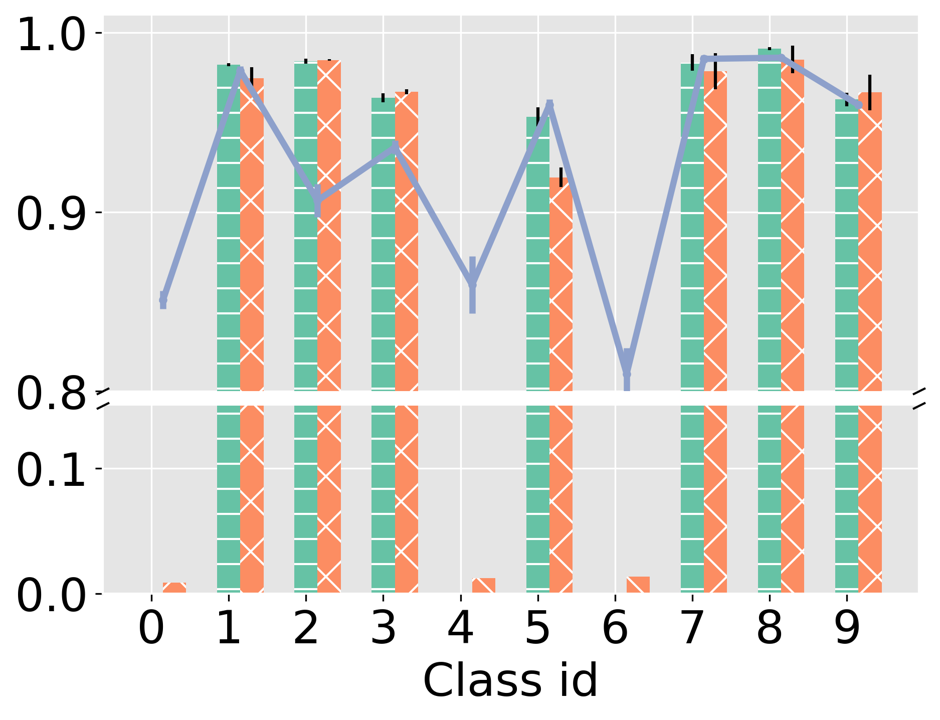}
        \caption{Unlearn $3$ classes} 
        \label{fig: cls-acc-fmnist-3}
    \end{subfigure}
    \hfill
         \begin{subfigure}[b]{0.24\textwidth}
        \includegraphics[width=\textwidth]{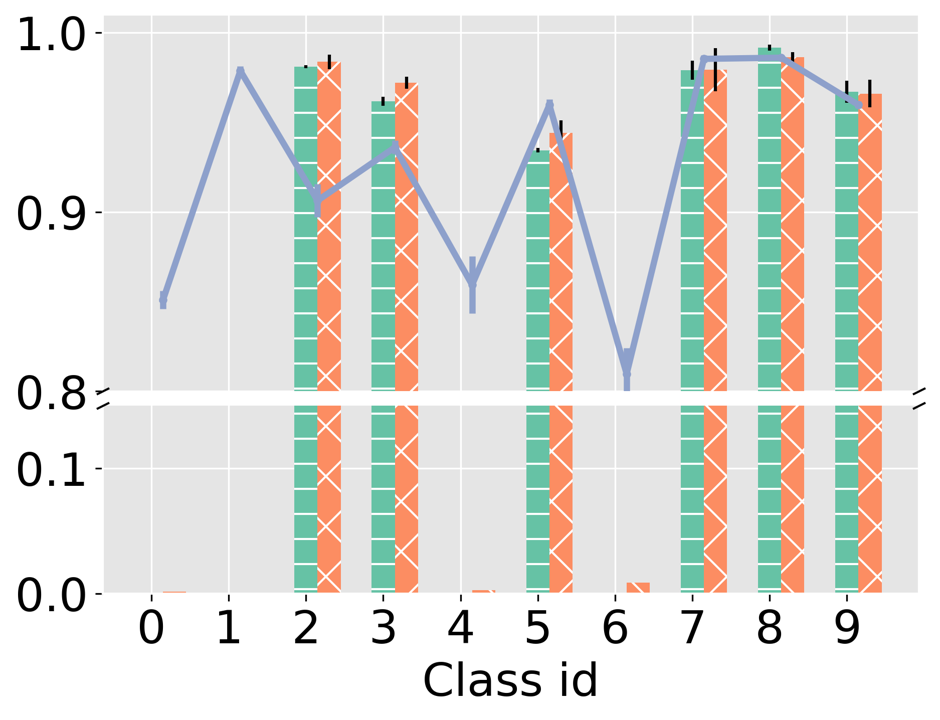}
        \caption{Unlearn $4$ classes} 
        \label{fig: cls-acc-fmnist-4}
    \end{subfigure}
    \hfill
         \begin{subfigure}[b]{0.24\textwidth}
        \includegraphics[width=\textwidth]{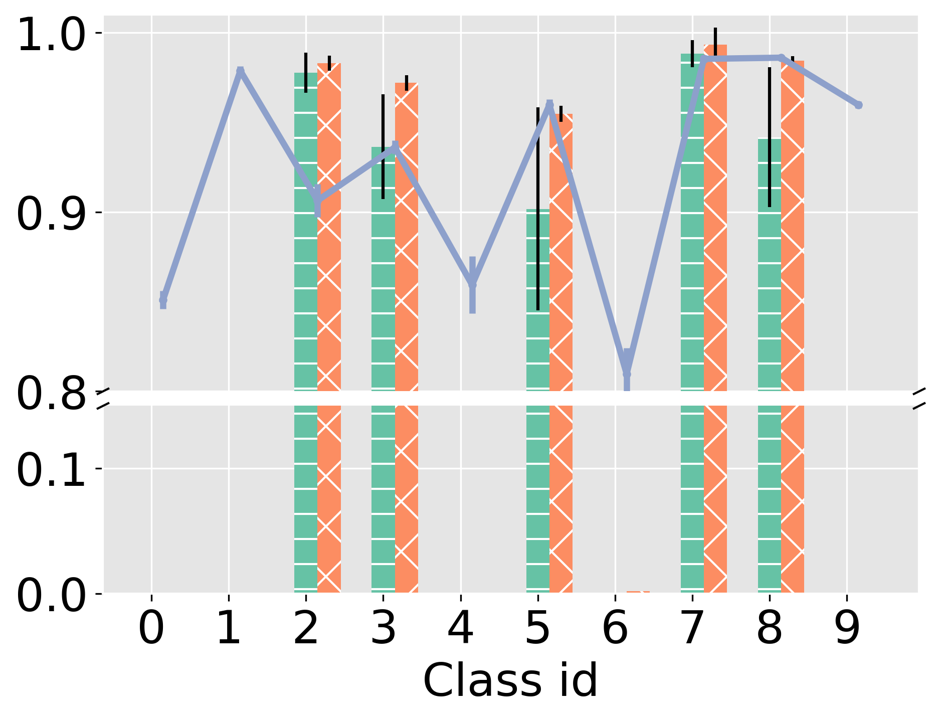}
        \caption{Unlearn $5$ classes} 
        \label{fig: cls-acc-fmnist-5}
    \end{subfigure}
    \caption{Class-wise accuracy of the original model, the retrained model, and the model obtained by UNSC on FashionMNIST. We present the results of unlearning different numbers of classes.}
    \label{fig: cls-acc-fmnist}
\end{figure*}

\begin{table*}[htp]
    \centering
    \caption{Performance evaluation of UNSC on FashionMNIST.} 
    \label{tab: multi-classes unlearn fmnist}
    \begin{tabular}{c|ccccccc}
         \toprule
         \multirow{2}{*}{\#Classes} & \multicolumn{2}{c}{Original model} & \multicolumn{2}{c}{Retrain} & \multicolumn{2}{c}{UNSC} & Accuracy Gain\\
         & $Acc_{\mathcal{D}_{rt}}$ & $Acc_{\mathcal{D}_{ut}}$ & $Acc_{\mathcal{D}_{rt}}$ & $Acc_{\mathcal{D}_{ut}}$ & $Acc_{\mathcal{D}_{rt}}(\uparrow)$ & $Acc_{\mathcal{D}_{ut}}(\downarrow)$ & $\Delta Acc_{\mathcal{D}_r}(\uparrow)$\\
         \midrule
         1 & $93.58 _{\pm 0.29} $ & $80.97 \pm1.44$ & $95.19 _{\pm 0.20} $ & $0.00 _{\pm 0.00}$ & $95.23 _{\pm 0.27} $ & $0.67 _{\pm 0.17}$ & $1.65\pm0.30$ \\
         2 & $94.64 _{\pm 0.35} $ & $83.03 _{\pm 0.54} $ & $96.16 _{\pm 0.04} $ & $0.00 _{\pm 0.00} $ & $95.38 _{\pm 0.38} $ & $0.80 _{\pm 0.29} $ & $0.74\pm0.26$\\
         3 & $95.88 _{\pm 0.22} $ & $84.00 _{\pm 0.33} $ & $97.42 _{\pm 0.08} $ & $0.00 _{\pm 0.00} $ & $96.80 _{\pm 0.29} $ & $1.16 _{\pm 0.59} $ & $0.91\pm0.21$\\
         4 & $95.55 _{\pm 0.22} $ & $87.47 _{\pm 0.31} $ & $96.91 _{\pm 0.02} $  & $0.00 _{\pm 0.00} $ & $97.19 _{\pm 0.21} $ & $0.35 _{\pm 0.41} $ & $1.64\pm0.21$\\
         5 & $95.47 _{\pm 0.26} $ & $89.17 _{\pm 0.26} $ & $94.91 _{\pm 1.93} $ & $0.00 _{\pm 0.00} $ & $97.88 _{\pm 0.15} $ & $0.04 _{\pm 0.02} $ & $2.41\pm0.24$\\
          \bottomrule
    \end{tabular}
\end{table*}

\begin{figure*}[htp]
    \centering
     \begin{subfigure}[b]{0.24\textwidth}
        \includegraphics[width=\textwidth]{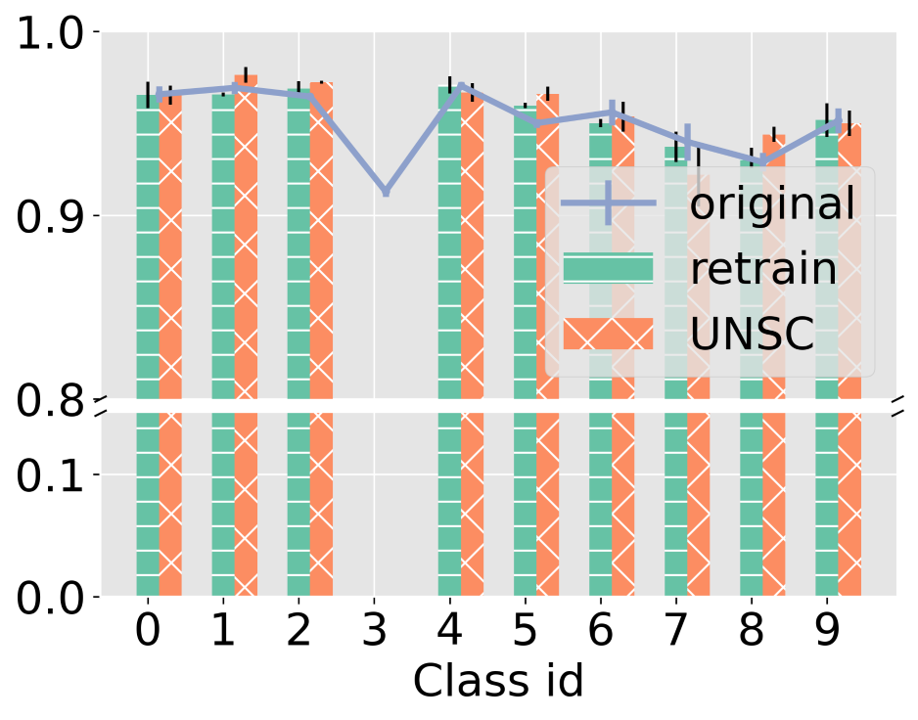}
        \caption{Unlearn $1$ class} 
        \label{fig: cls-acc-svhn-1}
    \end{subfigure}
    \hfill
         \begin{subfigure}[b]{0.24\textwidth}
        \includegraphics[width=\textwidth]{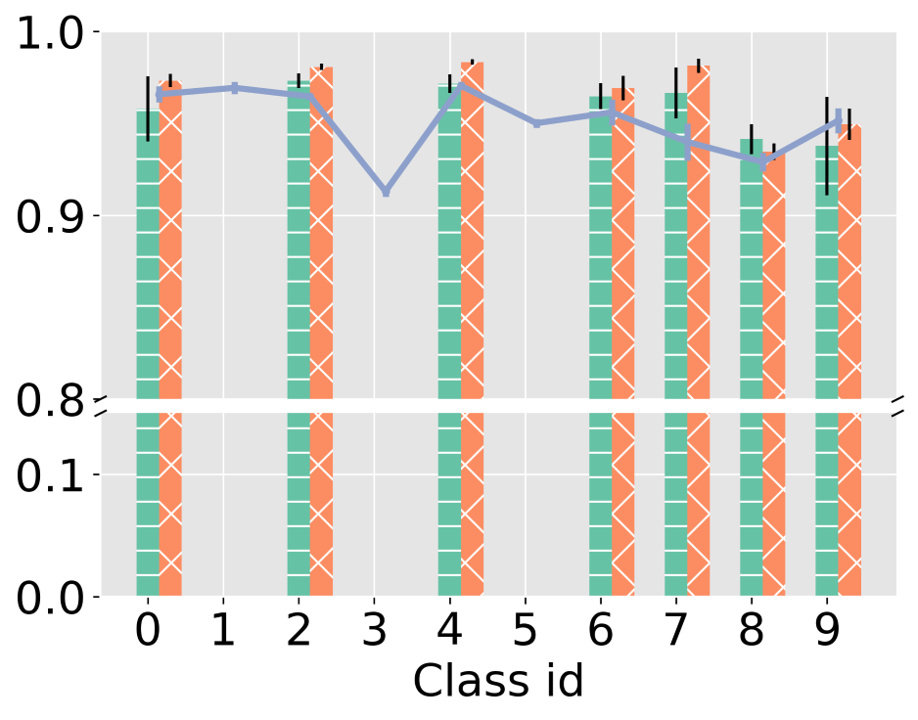}
        \caption{Unlearn $3$ classes} 
        \label{fig: cls-acc-svhn-3}
    \end{subfigure}
    \hfill
         \begin{subfigure}[b]{0.24\textwidth}
        \includegraphics[width=\textwidth]{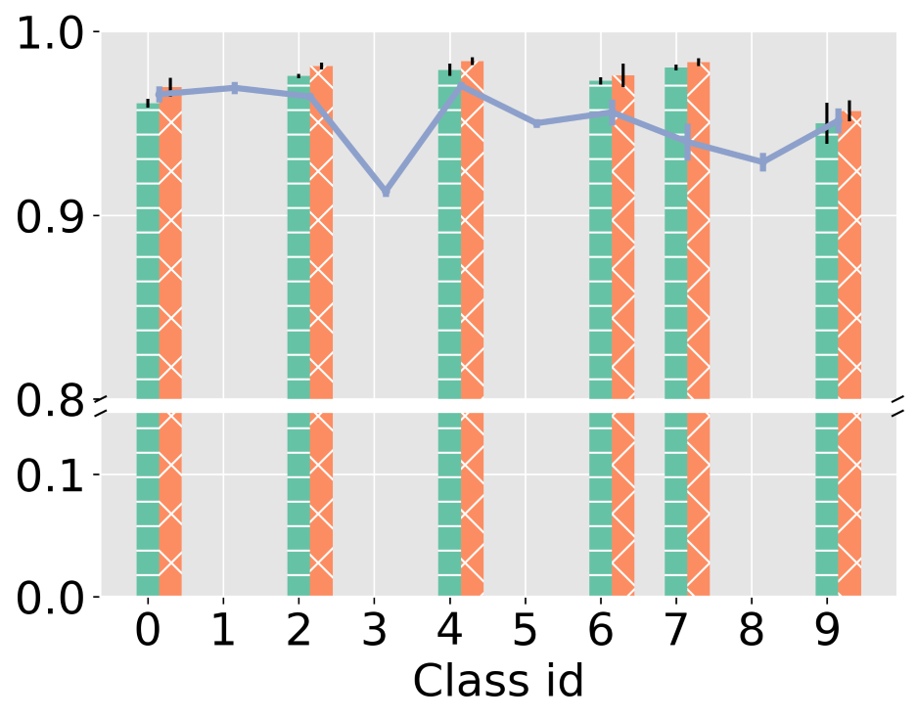}
        \caption{Unlearn $4$ classes} 
        \label{fig: cls-acc-svhn-4}
    \end{subfigure}
    \hfill
         \begin{subfigure}[b]{0.24\textwidth}
        \includegraphics[width=\textwidth]{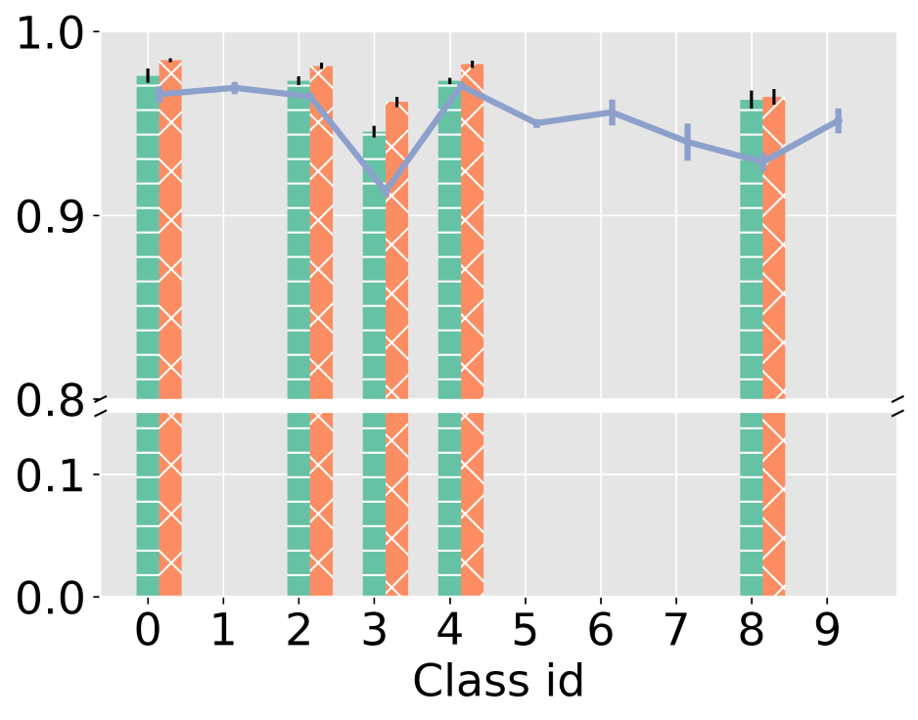}
        \caption{Unlearn $5$ classes} 
        \label{fig: cls-acc-svhn-5}
    \end{subfigure}
    \caption{Class-wise accuracy of the original model, the retrained model, and the model obtained by UNSC on SVHN. We present the results of unlearning different numbers of classes.}
    \label{fig: cls-acc-svhn}
\end{figure*}

\begin{table*}[htp]
    \centering
    \caption{Performance evaluation of UNSC on SVHN. } 
    \label{tab: multi-classes unlearn svhn}
    \begin{tabular}{c|ccccccc}
         \toprule
         \multirow{2}{*}{\#Classes} & \multicolumn{2}{c}{Original model} & \multicolumn{2}{c}{Retrain} & \multicolumn{2}{c}{UNSC} & Accuracy Gain\\
         & $Acc_{\mathcal{D}_{rt}}$ & $Acc_{\mathcal{D}_{ut}}$ & $Acc_{\mathcal{D}_{rt}}$ & $Acc_{\mathcal{D}_{ut}}$ & $Acc_{\mathcal{D}_{rt}}(\uparrow)$ & $Acc_{\mathcal{D}_{ut}}(\downarrow)$ & $\Delta Acc_{\mathcal{D}_r}(\uparrow)$\\
         \midrule
         1 & $95.52 _{\pm 0.12} $ & $91.30 _{\pm 0.30} $ & $95.56 _{\pm 0.23} $ & $0.00 _{\pm 0.00} $ & $95.74 _{\pm 0.17} $ & $0.00 _{\pm 0.00} $ & $0.22\pm0.09$\\
         2 &$95.34 _{\pm 0.14} $ & $94.11 _{\pm 0.30} $ & $96.33 _{\pm 0.08} $ & $0.00 _{\pm 0.00} $ & $96.44 _{\pm 0.15} $ & $0.00 _{\pm 0.00} $ & $1.10\pm0.09$\\
         3 & $95.39 _{\pm 0.15} $ & $94.41 _{\pm 0.13} $ & $95.90 _{\pm 1.15} $ & $0.00 _{\pm 0.00} $ & $96.74 _{\pm 0.16} $ & $0.01 _{\pm 0.02} $ & $1.35\pm0.11$\\
         4 & $95.80 _{\pm 0.20} $ & $94.03 _{\pm 0.05} $ & $96.99 _{\pm 0.13} $  & $0.00 _{\pm 0.00} $ & $97.51 _{\pm 0.10} $ & $0.01 _{\pm 0.01} $ & $1.70\pm0.12$\\
         5 & $94.85 _{\pm 0.08} $ & $95.34 _{\pm 0.22} $ & $96.61 _{\pm 0.17} $ & $0.00 _{\pm 0.00} $ & $97.48 _{\pm 0.04} $ & $0.00 _{\pm 0.00} $ & $2.62\pm0.13$\\
         \bottomrule
    \end{tabular}
\end{table*}

\begin{figure}
    \centering
    \scalebox{1.0}[0.85]{\includegraphics[width=\linewidth]{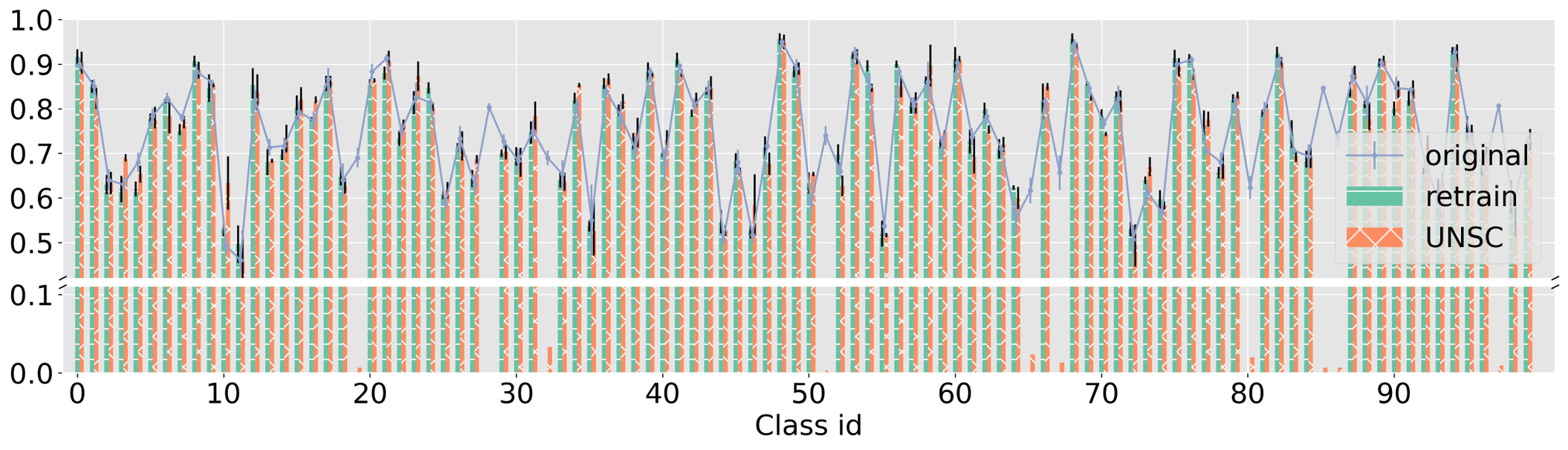}}
    \caption{Class-wise accuracy of the original model, the retrained model, and the model obtained by UNSC on CIFAR-100.We randomly unlearn 10 classes.}
    \label{fig: cls-acc-CIFAR-100}
\end{figure}

\begin{table*}[htp]
    \centering
    \caption{Performance evaluation of UNSC on CIFAR-100.} 
    \setlength{\tabcolsep}{3pt}
    \label{tab: multi-classes unlearn CIFAR-100}
    \begin{tabular}{c|ccccccc}
         \toprule
         \multirow{2}{*}{\#unlearn classes} & \multicolumn{2}{c}{Original model} & \multicolumn{2}{c}{Retrain} & \multicolumn{2}{c}{UNSC} & Accuracy Gain\\
         & $Acc_{\mathcal{D}_{rt}}$ & $Acc_{\mathcal{D}_{ut}}$ & $Acc_{\mathcal{D}_{rt}}$ & $Acc_{\mathcal{D}_{ut}}$ & $Acc_{\mathcal{D}_r} (\uparrow)$ & $Acc_{\mathcal{D}_{ut}}(\downarrow)$ & $\Delta Acc_{\mathcal{D}_r}(\uparrow)$\\
         \midrule
         10 &$75.36 _{\pm 0.34} $& $72.07 _{\pm 1.18} $ & $75.77 _{\pm 0.22} $& $0.00 _{\pm 0.00}$ & $76.16 _{\pm0.35} $ & $1.80_{\pm0.16} $ & $0.80\pm0.26$\\
         20 &$74.98 _{\pm 0.32} $ & $75.20 _{\pm 0.66} $ & $76.44 _{\pm 0.21} $ & $ 0.00 _{\pm 0.00} $ & $76.85 _{\pm 0.23} $ & $0.67 _{\pm 0.27} $ & $1.87 _{\pm 0.15} $ \\
         30 &$74.40 _{\pm 0.18} $ & $76.50 _{\pm 0.78} $ & $75.81 _{\pm 0.26} $ & $0.00 _{\pm 0.00} $ & $77.61 _{\pm 0.10} $ & $0.43 _{\pm 0.08} $ & $3.21 _{\pm 0.20} $\\
         40 & $73.94 _{\pm 0.27} $ & $76.65 _{\pm 0.52} $ & $76.77 _{\pm 0.12} $ & $0.00 _{\pm 0.00} $ & $78.52 _{\pm 0.22} $ & $0.31 _{\pm 0.13} $ & $4.58 _{\pm 0.32} $\\
         50 & $73.73 _{\pm 0.35} $ & $76.32 _{\pm 0.47} $ & $76.80 _{\pm 0.15} $ & $0.00 _{\pm 0.00} $ & $79.64 _{\pm 0.27} $ & $0.21 _{\pm 0.09} $ & $5.91 _{\pm 0.27}$\\
         \midrule
    \end{tabular}
\end{table*}

\newpage
\subsection{Loss Contours}
We provide loss contours of SVHN, FMNIST, and CIFAR-100 in Figure \ref{fig: loss contours on fmnist svhn and CIFAR-100}. Similar to Figure \ref{fig: loss contour}, on all datasets, loss contour shows the same pattern: loss remains approximately the same when altering the model's parameters in the null space. 
\begin{figure}[htp]
    \centering
    \begin{subfigure}[b]{0.3\columnwidth}
        \scalebox{1.0}[0.85]{\includegraphics[width=\textwidth]{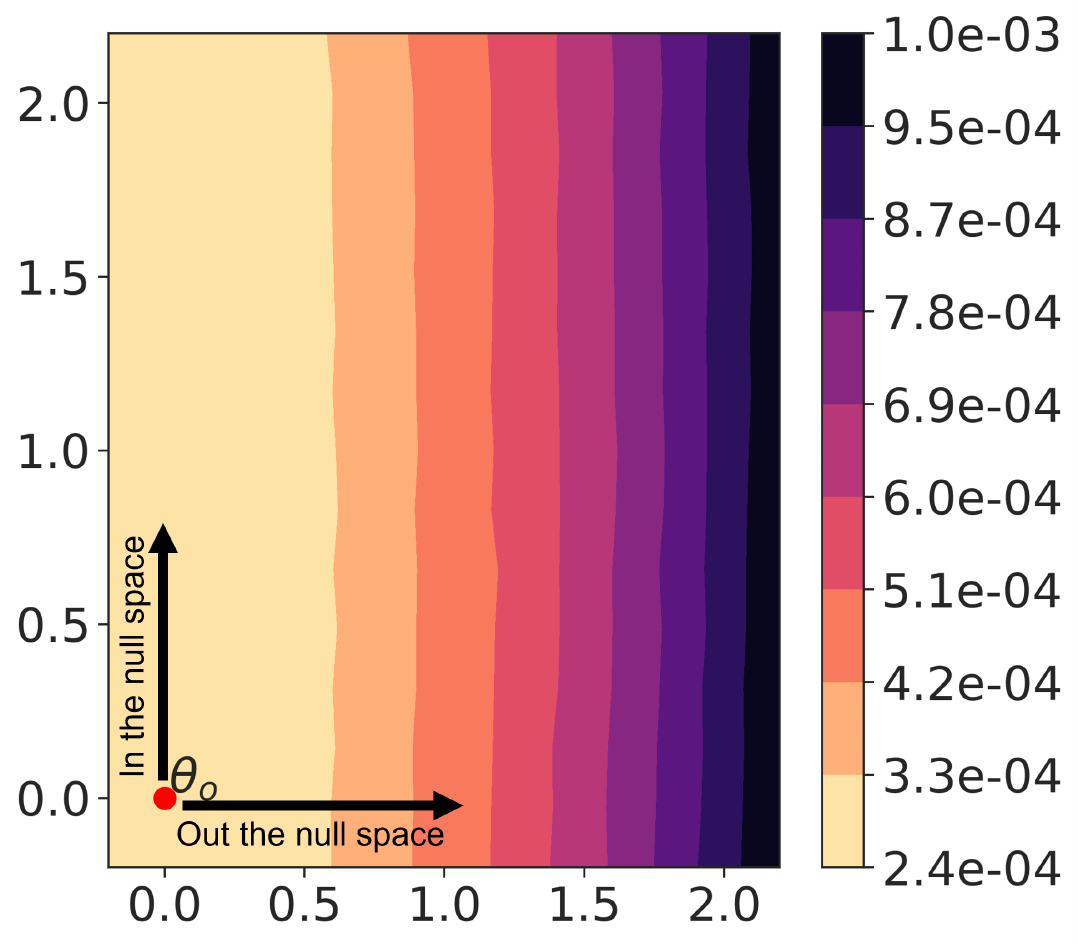}}
        \caption{FashionMNIST} 
    \end{subfigure}
    \hfill
    \begin{subfigure}[b]{0.3\columnwidth}
        \scalebox{1.0}[0.85]{\includegraphics[width=\textwidth]{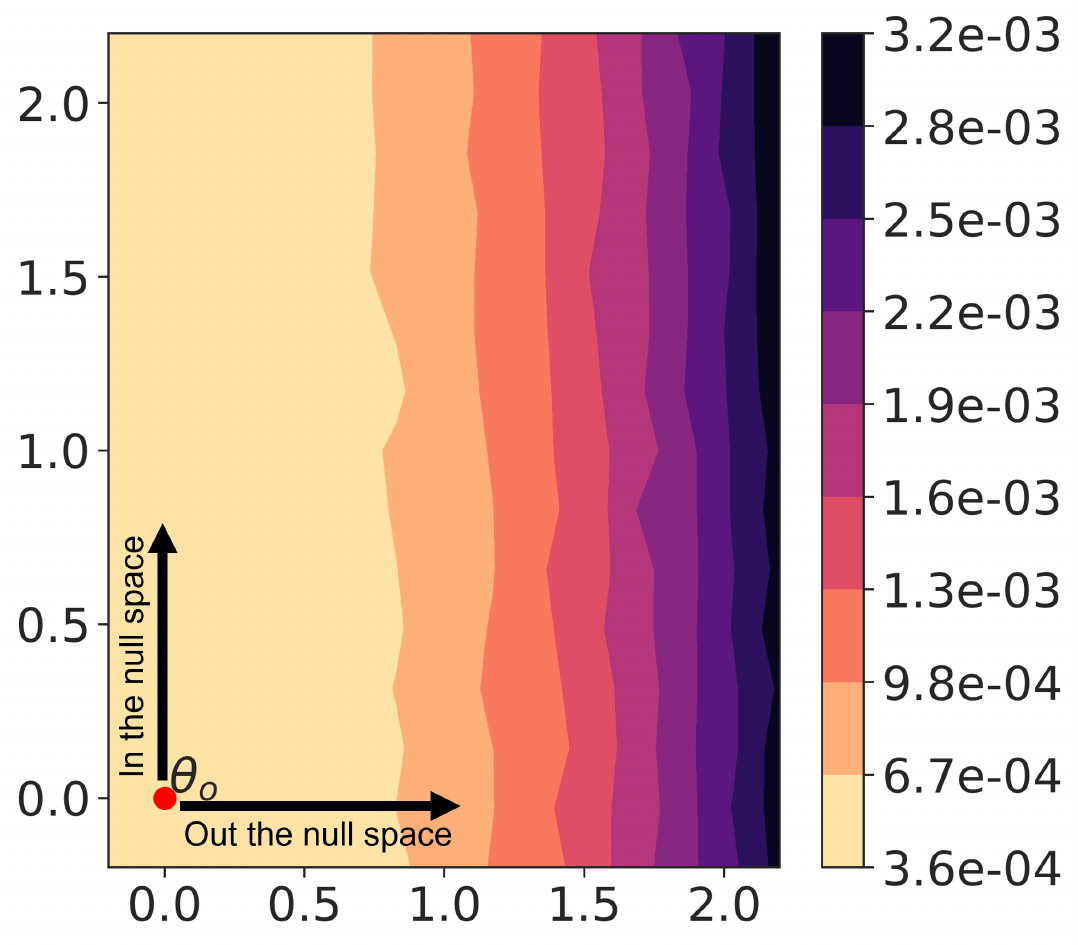}}
        \caption{SVHN} 
    \end{subfigure}
    \hfill
    \begin{subfigure}[b]{0.3\columnwidth}
        \scalebox{1.0}[0.85]{\includegraphics[width=\textwidth]{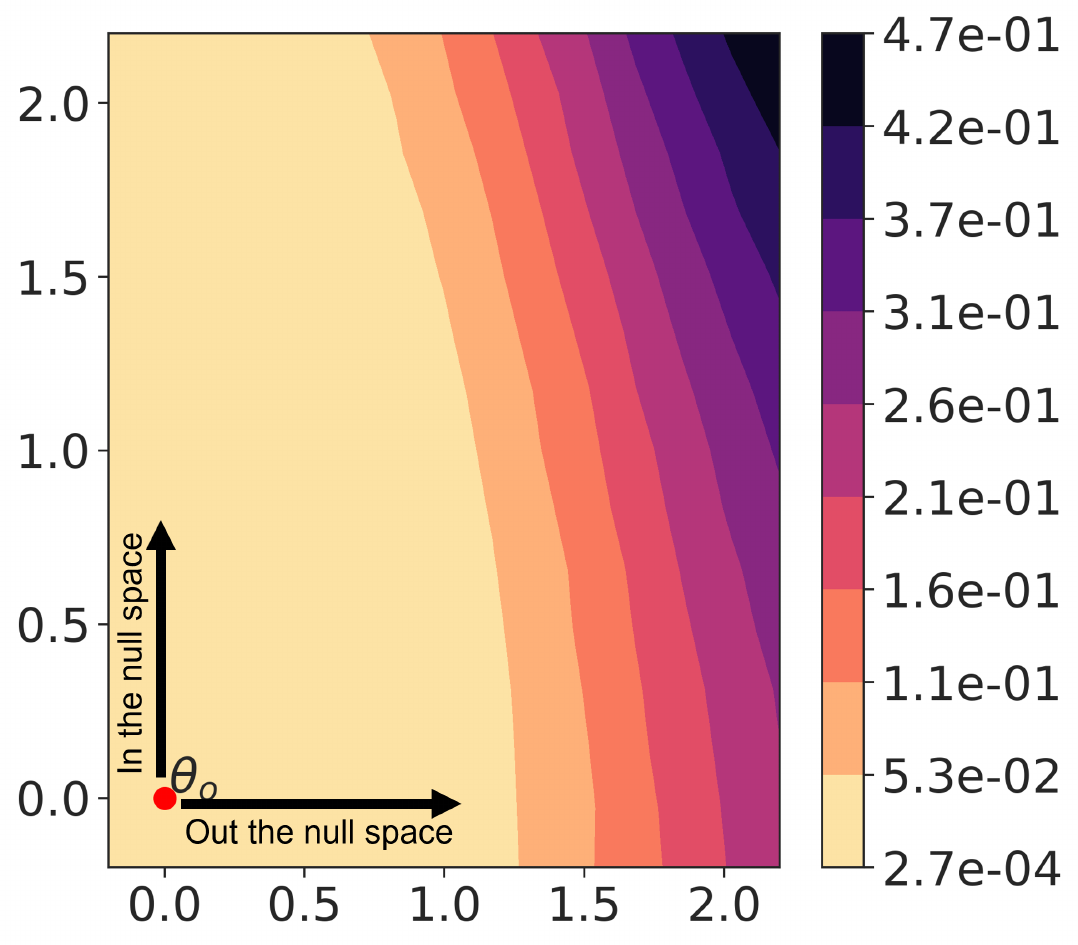}}
        \caption{CIFAR-100} 
    \end{subfigure}
    \caption{Loss contours on FashionMNIST, SVHN, and CIFAR-100}
    \label{fig: loss contours on fmnist svhn and CIFAR-100}
\end{figure}
\vspace{-1em}

\subsection{Distributions of Pseudo-lables}
We compare the distributions of pseudo-labels on FashionMNIST, SVHN, and CIFAR-100 datasets. On all datasets, the assigned pseudo-labels closely align with the predictions given by the retrained model.
\begin{figure*}[htp]
    \centering
    \begin{subfigure}[b]{0.165\textwidth}
        \includegraphics[width=\textwidth]{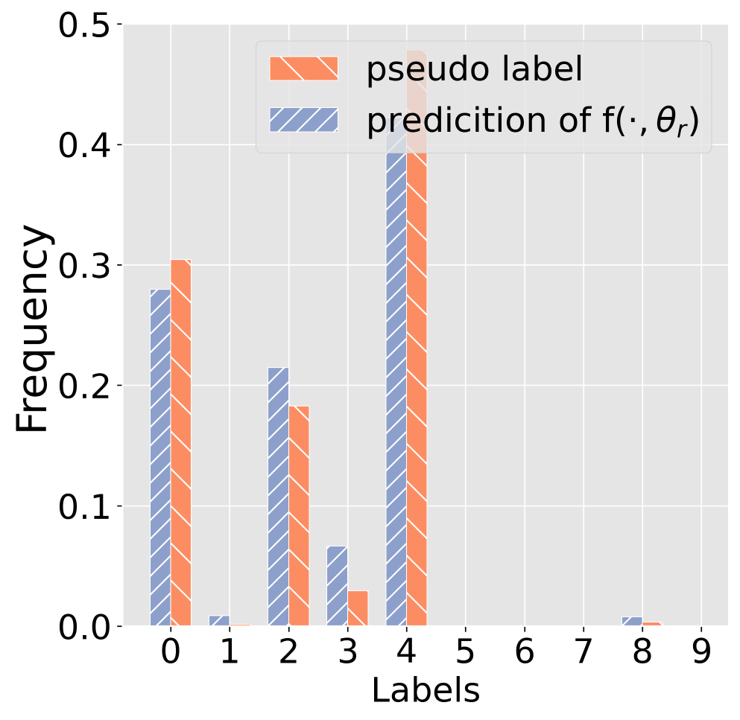}
        \caption{FashionMNIST} 
    \end{subfigure}
    \begin{subfigure}[b]{0.165\textwidth}
        \includegraphics[width=\textwidth]{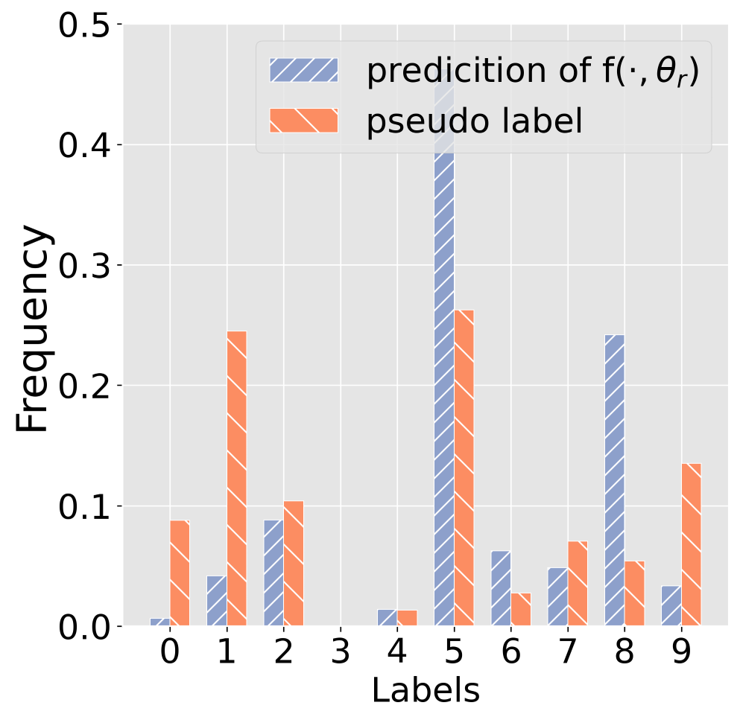}
        \caption{SVHN} 
    \end{subfigure}
    \begin{subfigure}[b]{0.58\textwidth}
    \centering
        \includegraphics[width=\textwidth]{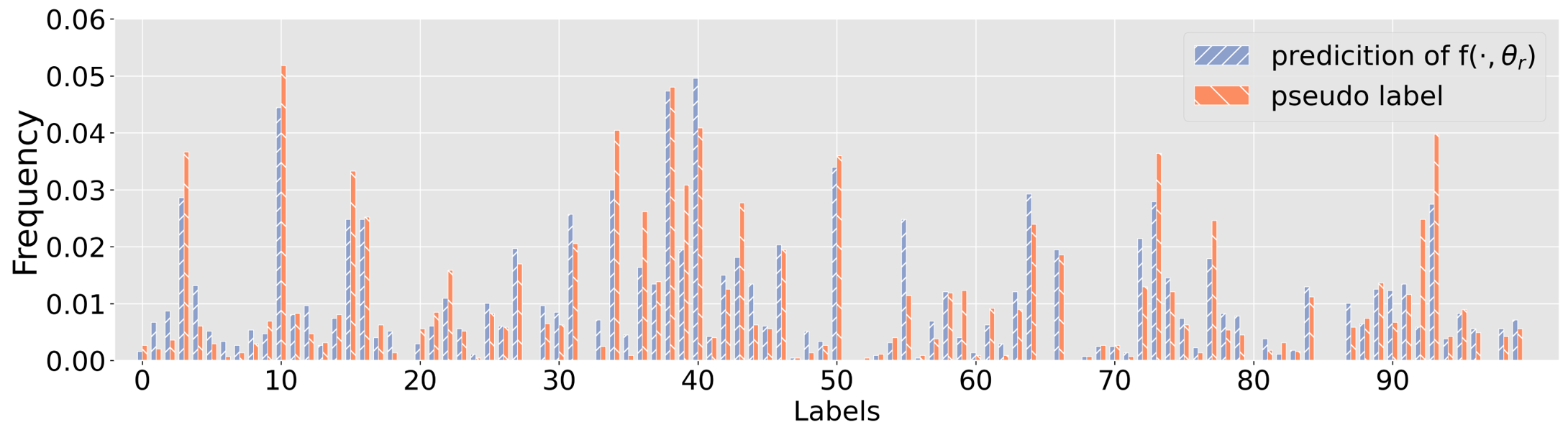}
        \caption{CIFAR-100} 
    \end{subfigure}
    \caption{Distributions of pseudo labels and labels assigned by the retrained model: (a) FashionMNIST, (b) SVHN, (c) CIFAR-100.}
\end{figure*}
\vspace{-1em}

\subsection{Ablation Studies on SVHN, CIFAR-10, and CIFAR-100}
Following the same setting in Section \ref{sec: exp}, we randomly select one class on SVHN and CIFAR-10 (ten classes on CIFAR-100) and unlearn all the samples. Tables \ref{tab: ablation svhn}, \ref{tab: ablation CIFAR-10}, and \ref{tab: ablation CIFAR-100} present ablation results.

\begin{table}[htp]
    \centering
    \caption{Ablation results on SVHN.}
    \vspace{-0.5em}
    \label{tab: ablation svhn}
    \begin{tabular}{c|ccccc|cc}
        \toprule
         Exp. ID & \begin{tabular}[c]{@{}c@{}}Original \\ model\end{tabular} & \begin{tabular}[c]{@{}c@{}}Retrain\\model\end{tabular} & RL
         &  \begin{tabular}[c]{@{}c@{}}Null \\space\end{tabular} & \begin{tabular}[c]{@{}c@{}}Pesudo-\\labeling\end{tabular} & $Acc_{\mathcal{D}_{rt}}$ & $Acc_{\mathcal{D}_{ut}}$\\
         \midrule
         \midrule
         I & $\checkmark$ & & & & & $95.52 _{\pm 0.12}$ & $91.30 _{\pm 0.30}$\\
         II & & $\checkmark$ & & & & $95.56 _{\pm 0.23}$ & $0.00_{\pm 0.00} $\\
         III & & &  $\checkmark$ & & & $75.87_{\pm 8.67}$ & $1.17_{\pm 0.16} $\\
         \midrule
         IV & & & $\checkmark$ & $\checkmark$ & & $93.64 _{\pm 0.71}$ & $2.96 _{\pm 2.11} $\\
         V & & & & $\checkmark$ & $\checkmark$& $95.74 _{\pm 0.17}$ & $0.00 _{\pm 0.00}$\\
         \bottomrule
    \end{tabular}
\end{table}
\vspace{-1.5em}
\begin{table}[H]
    \centering
    \caption{Ablation results on CIFAR-10.}
    \vspace{-0.5em}
    \label{tab: ablation CIFAR-10}
    \begin{tabular}{c|ccccc|cc}
        \toprule
         Exp. ID & \begin{tabular}[c]{@{}c@{}}Original \\ model\end{tabular} & \begin{tabular}[c]{@{}c@{}}Retrain\\model\end{tabular} & RL
         &  \begin{tabular}[c]{@{}c@{}}Null \\space\end{tabular} & \begin{tabular}[c]{@{}c@{}}Pesudo-\\labeling\end{tabular} & $Acc_{\mathcal{D}_{rt}}$ & $Acc_{\mathcal{D}_{ut}}$\\
         \midrule
         \midrule
         I & $\checkmark$ & & & & & $92.10_{\pm0.29}$ & $79.93_{\pm1.72}$\\
         II & & $\checkmark$ & & & & $93.22_{\pm0.18}$ & $0.00_{\pm0.00}$\\
         III & & & $\checkmark$ & & & $82.88_{\pm 2.62}$ & $3.30_{\pm0.83}$\\
         \midrule
         IV & &  & $\checkmark$ & $\checkmark$ & & $92.71 _{\pm 0.31} $ & $2.10 _{\pm 0.33} $\\
         V & & &  & $\checkmark$ & $\checkmark$& $93.09_{\pm 0.25} $ & $1.30_{\pm0.59}$\\
         \bottomrule
    \end{tabular}
\end{table}
\vspace{-1.5em}
\begin{table}[H]
    \centering
    \caption{Ablation results on CIFAR-100.}
    \vspace{-0.5em}
    \label{tab: ablation CIFAR-100}
    \begin{tabular}{c|ccccc|cc}
        \toprule
         Exp. ID & \begin{tabular}[c]{@{}c@{}}Original \\ model\end{tabular} & \begin{tabular}[c]{@{}c@{}}Retrain\\model\end{tabular} & RL
         &  \begin{tabular}[c]{@{}c@{}}Null \\space\end{tabular} & \begin{tabular}[c]{@{}c@{}}Pesudo-\\labeling\end{tabular} & $Acc_{\mathcal{D}_{rt}}$ & $Acc_{\mathcal{D}_{ut}}$\\
         \midrule
         \midrule
         I & $\checkmark$ & & & & & $75.36 _{\pm 0.34}$ & $72.07 _{\pm 1.18}$\\
         II & & $\checkmark$ & & & & $75.77 _{\pm 0.22}$ & $0.00_{\pm0.00}$\\
         III & & & $\checkmark$ & & & $37.63_{\pm 1.98}$ & $5.27_{\pm0.48}$\\
         \midrule
         IV & & & $\checkmark$ & $\checkmark$ & & $73.41 _{\pm 0.47}$ & $0.30 _{\pm 0.22}$\\
         V & & &  & $\checkmark$ & $\checkmark$& $76.16 _{\pm0.35}$ & $1.80_{\pm0.16}$\\
         \bottomrule
    \end{tabular}
\end{table}
\vspace{-5em}
\end{nolinenumbers}
\end{document}